\pdfoutput=1
\documentclass[11pt,a4paper]{article}
\usepackage{hyperref}
\usepackage[hyperref]{emnlp2020}
\usepackage{times}
\usepackage{graphicx}
\usepackage{latexsym}
\usepackage{enumitem}
\usepackage{xcolor}
\usepackage{amsmath}
\usepackage{subfig}
\usepackage{array}
\usepackage{tabu}

\usepackage{microtype}

\aclfinalcopy % Uncomment this line for the final submission
\DeclareMathOperator*{\argmax}{arg\,max}

\title{memeBot: Towards Automatic Image Meme Generation}

\author{}

\author{Aadhavan Sadasivam,  Kausic Gunasekar,  Hasan Davulcu, Yezhou Yang \\
        Arizona State University, Tempe AZ, United States \\ 
        \{asadasi1, kgunase3, hdavulcu, yz.yang \}@asu.edu}

\date{}

\begin{document}
\maketitle
\begin{abstract}
Image memes have become a widespread tool used by people for interacting and exchanging ideas over social media, blogs, and open messengers. This work proposes to treat automatic image meme generation as a translation process, and further present an end to end neural and probabilistic approach to generate an image-based meme for any given sentence using an encoder-decoder architecture. For a given input sentence, an image meme is generated by combining a meme template image and a text caption where the meme template image is selected from a set of popular candidates using a selection module, and the meme caption is generated by an encoder-decoder model. An encoder is used to map the selected meme template and the input sentence into a meme embedding and a decoder is used to decode the meme caption from the meme embedding. The generated natural language meme caption is conditioned on the input sentence and the selected meme template. The model learns the dependencies between the meme captions and the meme template images and generates new memes using the learned dependencies. The quality of the generated captions and the generated memes is evaluated through both automated and human evaluation. An experiment is designed to score how well the generated memes can represent the tweets from Twitter conversations. Experiments on Twitter data show the efficacy of the model in generating memes for sentences in online social interaction.
\end{abstract}

\section{Introduction}

An Internet meme commonly takes the form of an image and is created by combining a meme template (image) and a caption (natural language sentence). The image typically comes from a set of popular image candidates and the caption conveys the intended message through natural language.

\begin{figure}[t]
    \centering
    \includegraphics[width=1.0\columnwidth]{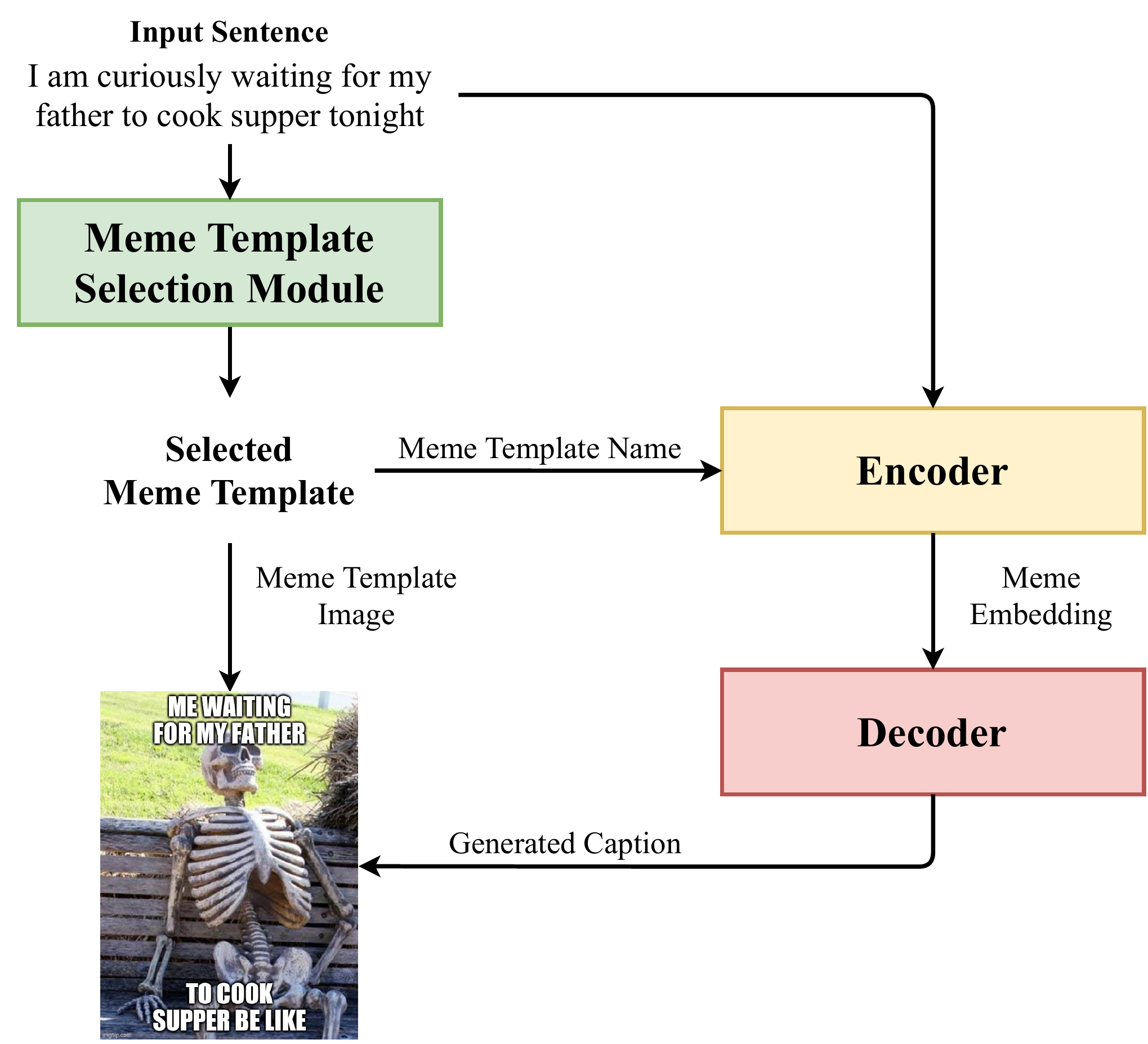}
    \caption{An illustrative figure of memeBot. It generates an image meme for a given input sentence by combining the selected meme template image and the generated meme caption.}
    \label{fig:illustrative_figure}
\end{figure}

\begin{figure}[h]
    \centering
    \includegraphics[width=\columnwidth]{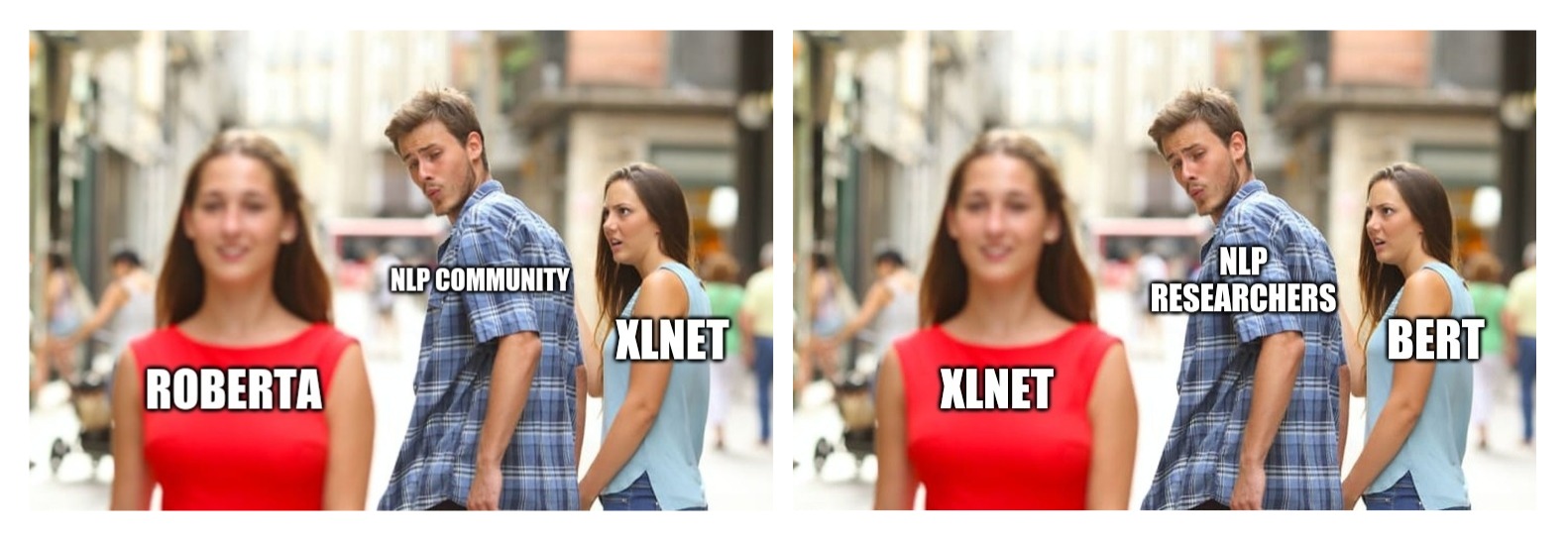}
    \caption{Memes used by the online deep learning community on social media to ridicule the state-of-the-art pre-training models.}
    \label{fig:sample}
\end{figure}

Over the internet, information exists in the form of text, images, video, or a combination of these. However the information existing as a combination of image or video and short text often gets viral. Image memes are a combination of image, text, and humor, making them a powerful tool to deliver information. The image memes are popular because they portray the culture and social choices embraced by the internet community and they have a strong influence on the cultural norms of how specific demographics of people operate. For example, in Figure ~\ref{fig:sample}, we present the memes used by an online deep learning community to ridicule how the new pre-training methods are outperforming the previous state-of-the-art models.

The popularity of image-based memes can be attributed to the fact that visual information is easier to process and understand when compared to reading large blocks of text, and this fact is evident in Figure ~\ref{fig:sample}. The key role played by the image memes in shaping the popular culture of the internet community makes automatic meme image generation a demanding research topic to delve into.

\citet{davison2012language} separates a meme into three components - Manifestation, Behavior, and Ideal. In an image meme, the Ideal is the idea that needs to be conveyed. The Behavior is to select a suitable meme template and caption to convey that idea and, the Manifestation is the final meme image with a caption conveying the idea. \citet{wang2015can} and \citet{peirson2018dank} focus on the behavior and manifestation of a meme, not much importance is given to the ideal of a meme. Their approach of image meme generation is limited to selecting the most appropriate meme caption or generating a meme caption for the given image and template name. In this work, we intend to automatically generate an image meme to represent a given input sentence (Ideal) as illustrated in Figure 1, which is a challenging NLP task with immediate practical applications for online social interaction.

By taking a deep look into the process of image meme generation, we propose to co-relate image meme generation to Natural Language Translation. To translate a sentence from a source to target language, one has to decode the meaning of the sentence in its entirety, analyze its meaning, and then encode that meaning of the source sentence into the target sentence. Similarly, a sentence can be translated into a meme by encoding the meaning of the sentence into a pair of image and caption capable of conveying the same meaning or emotion as that of the sentence. Motivated by this intuition for meme generation, we develop a model that operates beyond the known approaches and extend the capability of image meme generation to generate memes for a given input sentence. We summarize our contributions as follows: 

\begin{itemize}
    \item We present an end to end encoder-decoder architecture to generate an image meme for any given sentence.
    \item We compiled the first large-scale Meme Caption dataset. %Upon publication, we will make the dataset and our implementation publicly available for the research community to conduct further investigation.
    \item We design experiments based on human evaluation and provide a thorough analysis of the experimental results on using the generated memes for online social interaction.
\end{itemize}

%Model architecture
\begin{figure*}[ht!]
    \centering
    \includegraphics[width=1.0\textwidth]{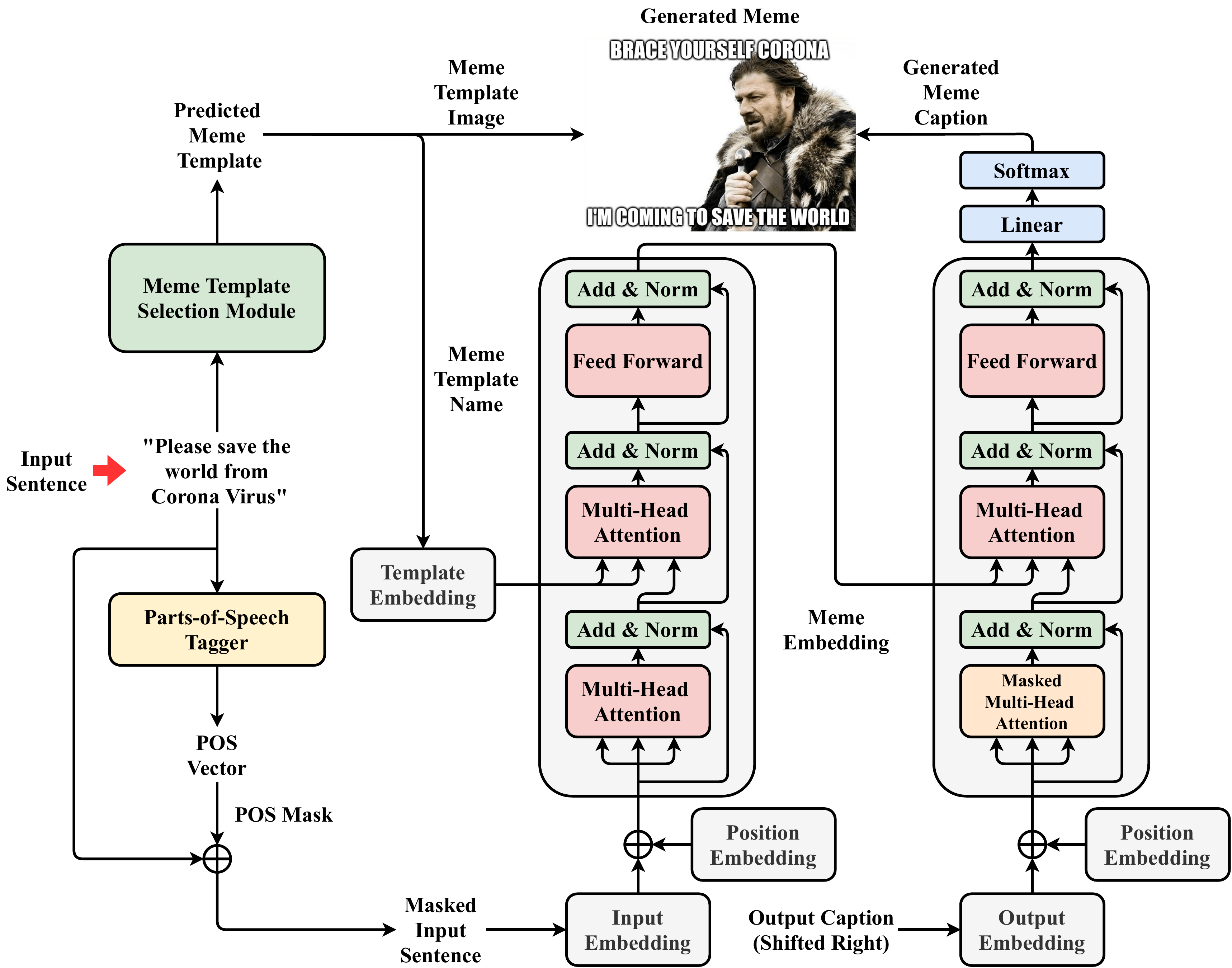}
    \caption{memeBot - model architecture. For a given input sentence, a meme is created by combining the meme image selected by the template selection module and the meme caption generated by the caption generation transformer.}
    \label{fig:architecture}
\end{figure*}

\section{Related Work}
There are only a few studies on automatic image meme generation and the existing approaches treat meme generation as a caption selection or caption generation problem. \citet{wang2015can} combined an image and its text description to select a meme caption from a corpus based on a ranking algorithm. \citet{peirson2018dank} extends Natural Language Description Generation to generating a meme caption using an encoder-decoder model with an attention \citep{luong-etal-2015-effective} mechanism. Although there is not much work on automatic meme generation, the task of meme generation can be closely aligned with tasks like Sentiment Analysis, Neural Machine Translation, Image Caption Generation and Controlled Natural Language Generation.

In Natural Language Understanding (NLU), researchers have explored classifying a sentence based on their sentiment \citep{socher2013recursive}. We extend this idea to classify a sentence based on its compatibility with a meme template. The idea of creating an encoded representation and decoding it into a desired target is well establish in Neural Machine Translation and Image Caption Generation. \citet{sutskever2014sequence}, \citet{nmt} and \citet{vaswani2017attention} use an encoder-decoder model to encode and decode a sentence from a source to a targeted language. \citet{showtell}, \citet{showattendtell}, \citet{imgdesc}  encode the visual features of an image and use a decoder to generate natural language description of the image. \citet{fang2020video2commonsense} generate natural language description containing common sense knowledge from the encoded visual inputs.

Our proposed model of meme generation shares similar spirit with the above mentioned problems where we encode the given input sentence into a latent space followed by decoding it into a meme caption that can be combined with the meme image to convey the same meaning as that of the input sentence. However, the generated meme caption should represent the input sentence through the selected meme template, making it a conditioned or controlled text generation task. Controlled Natural Language Generation with desired emotions, semantics and keywords have been studied previously. \citet{huang2018automatic} generate text with desired emotions by embedding emotion representations into Seq2Seq models. \citet{hu2017toward} concatenate a control vector to the latent space of their model to generate text with designated semantics. \citet{su2018incorporating} and \citet{miao2019cgmh} generate a sentence with desired emotions or keywords using sampling techniques. While these approaches of controlled text generation involve relatively complex conditioning factors, we implement a transformer \citep{vaswani2017attention} based encoder-decoder model to generate a meme caption conditioned on both the input sentence and the selected meme template.

\section{Our Approach}

In this section, we describe our approach: an end-to-end neural and probabilistic architecture for meme generation. Our model has two components. First, a meme template selection module to identify a compatible meme template (image) for the input sentence. Second, a meme caption generator as illustrated in Figure \ref{fig:architecture}.

\subsection{Meme Template Selection Module}

 Pre-trained language representations from transformer based architectures like BERT \cite{devlin-etal-2019-bert}, XLNet \cite{yang2019xlnet} and Roberta \citep{liu2019roberta} are being used in a wide range of Natural Language Understanding tasks. \citet{devlin-etal-2019-bert}, \citet{yang2019xlnet} and \citet{liu2019roberta} show that these models can be fine-tuned specifically to a range of NLU tasks to create state-of-the-art models. 

 For meme template selection module, we fine tune the pre-trained language representation models with a linear neural network on the meme template selection task. In training, the probability of selecting the correct template for a given sentence is maximized by using the formulation given below:
\begin{equation}
    l(\theta_{1}) = \argmax_{\theta_1} \sum_{(T,S)}log(P(T|S, \theta_1)),
\end{equation}
where $\theta_1$ denotes the parameters of the meme template selection module, $T$ is the template and $S$ is the input sentence.

\subsection{Meme Caption Generator}

We train the meme caption generator by corrupting the input caption, borrowing from denoising autoencoder \citep{vincent2008extracting}. We extract the parts of speech of the input caption using a Part-Of-Speech Tagger (POS Tagger) \citep{spacy2}. Using the POS vector, we mask the input caption such that only the noun phrases and verbs are passed as input to the meme caption generator. We corrupt the data to facilitate our model to learn meme generation from existing captions and to generalize the process of meme generation for any given input sentences during inference.

The meme cation generator model uses a transformer architecture inspired from \citet{vaswani2017attention}. Our transformer encoder creates meme embedding for a given sentence by performing multi-head scaled dot-product attention on the selected meme template and the input sentence. The transformer decoder initially performs masked multi-head attention on the expected caption and later performs multi-head scaled dot-product attention between the encoded meme embedding and the outputs of the masked multi-head attention as shown in Figure \ref{fig:architecture}. This enables the meme caption generator to learn the dependency between the input sentence, selected meme template and the expected meme caption. We optimize the transformer by using the formulation given below:

\begin{equation}
    l(\theta_2) = \argmax_{\theta_2} \sum_{(S,C)}log(P(C|M, \theta_2)),
\end{equation}
where $\theta_2$ denotes the parameters of the meme caption generator, $C$ is the meme caption and $M$ is the meme embedding obtained from the transformer encoder.

\section{Dataset}

\subsection{Meme Caption Dataset}\label{sec:data}

\begin{table*}[htbp]
    \begin{center}
     \begin{tabu}{ p{3cm}   m{8cm}  p{3cm}  }
        \tabucline[1pt]-
        \centering
        \textbf{Template Name} 
        &
        \centering
        \textbf{Captions} 
        &
        \textbf{Template Image}
        \\\tabucline[1pt]-
        \centering
        Leonardo Dicaprio Cheers
        &
        \begin{itemize}[noitemsep]
            \item to those who have been fortunate enough to have known true love
            \item cheers to us making the future bright
            \item when you see your cousin at a family gathering
        \end{itemize}
        &
        \begin{minipage}{3cm}
            \begin{center}
                \includegraphics[width= 2cm]{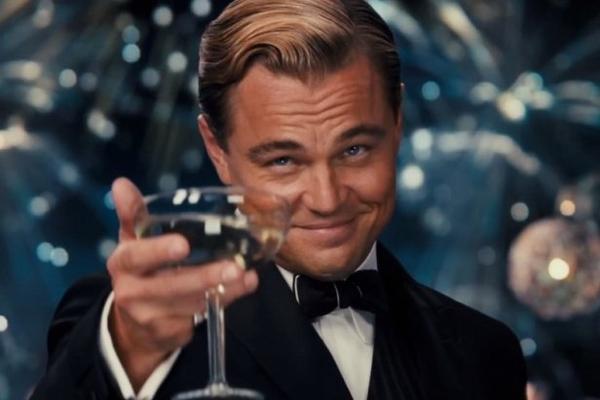}
            \end{center}
        \end{minipage} 
        \\ \hline
        \centering
        Success Kid
        &
        \begin{itemize}[noitemsep]
            \item carries the laundry didn't drop a single sock
            \item when you win your first fortnite game
            \item late to work and boss was even later
            \item when she gives you her phone number
        \end{itemize}
        &
        \begin{minipage}{3cm}
            \begin{center}
                \includegraphics[width= 2cm]{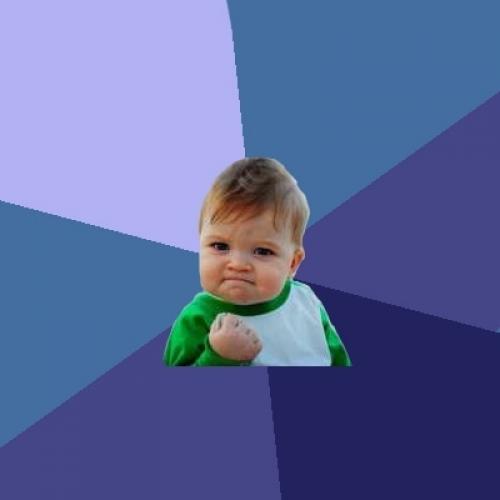}
            \end{center}
        \end{minipage}
        \\ \tabucline[1pt]-
        \end{tabu}
        \caption{Sample examples (Template name, Captions and Meme Image) from the meme caption dataset.}
      \label{tbl:sample_dataset}
    \end{center}
\end{table*}

%caption distribution
\begin{figure*}[h]
    \centering
    \includegraphics[width=1.0\textwidth]{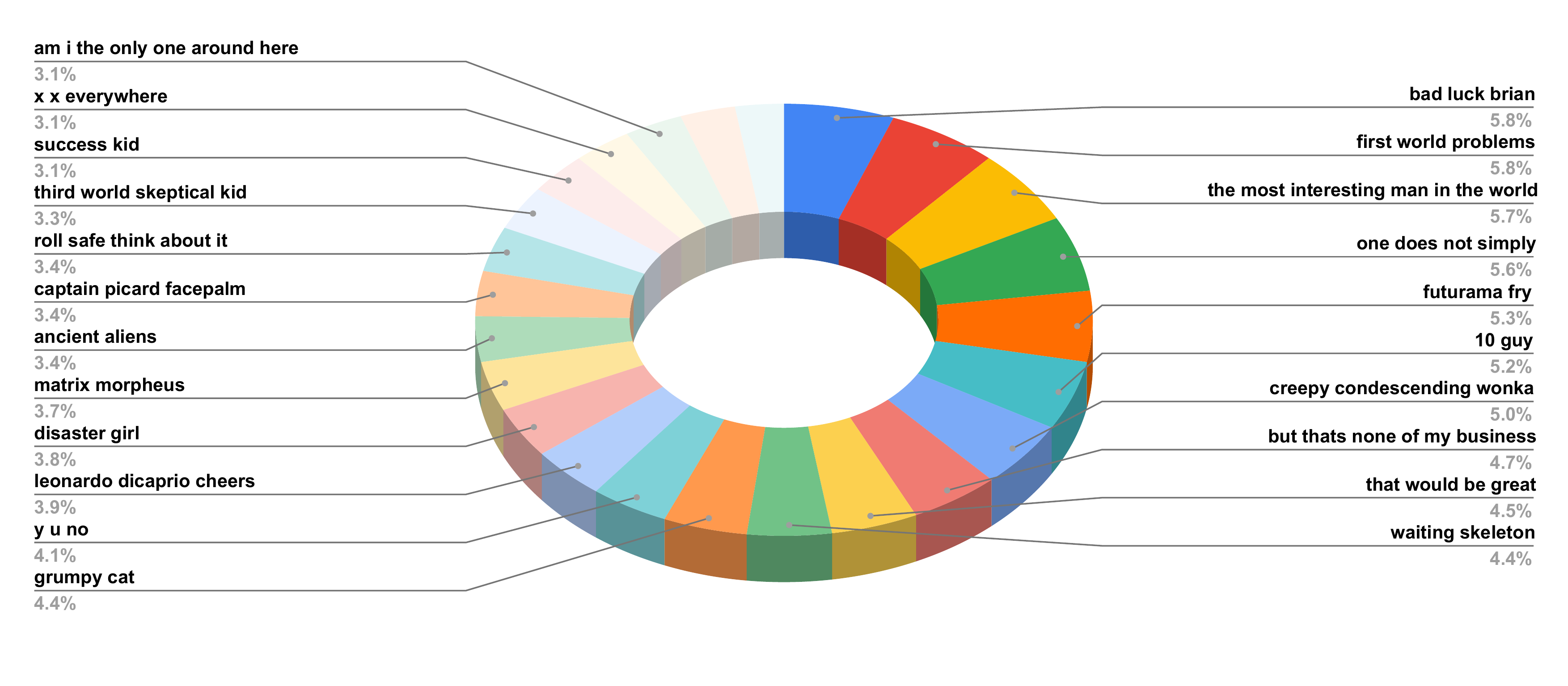}
    \caption{Distribution of meme caption count for the meme templates.}
    \label{fig:caption_distribution}
\end{figure*}

To make possible and validate the aforementioned technical framework, we collect a dataset that would enable us to learn the dependency between a meme template and a meme caption. Here we adopt the open online resource imgflip\footnote{\color{darkblue}{\href{https://imgflip.com}{https://imgflip.com}}} which is one of the most commonly used meme generators. To automatically crawl the data, we developed a web crawler to collect the memes. 

We observe that only a few meme templates dominate the collection. We investigate this dominating memes along with factors that can make a meme popular. Replication of a meme depends on the mental processes of observation and learning of the group of people across which it is being shared \cite{davison2012language}. Popular meme templates make a content shareable and are replicated frequently because of their capability to a make content viral. To this end, we experiment on image meme generation using the popular meme templates.

 Our dataset has $177,942$ meme captions from 24 templates. The distribution of meme captions across the meme templates is presented in Figure ~\ref{fig:caption_distribution}. The dataset consists of meme template (image $\&$ template name) and meme caption pairs. A sample from the dataset is illustrated in Table ~\ref{tbl:sample_dataset}. To add diversity to the generated memes, we use various images for the same meme template. The meme template figures used and a sample of the additional images used for the meme templates are presented in Appendices A and B.

\subsection{Twitter Dataset}

We collect tweets from Twitter to evaluate the efficacy of our model in generating memes for sentences used in online social interaction. We randomly sampled 6000 tweets for the query ``\textbf{Corona virus}". We prune the sampled tweets to 1000 tweets by selecting only those tweets with non negative sentiment using VADER-Sentiment-Analysis \citep{hutto2014vader}. Twitter is an open domain and may contain tweets that could affect the beliefs and sentiments of people and to have a control over our model, we remove the tweets with negative sentiments. The goal is to prompt our model to generate an image meme by inputting a tweet and evaluate if the generated meme is relevant to the tweet. 

\section{Experiments and Results}

We train our model on the meme caption dataset (Section ~\ref{sec:data}). The train, validation and test dataset contains $142341$, $17802$ and $17799$ samples respectively. We evaluate the performance of the meme template selection module in selecting the compatible template, the effectiveness of the caption generator in generating captions that are similar to the input captions from the meme caption test dataset and the efficacy of the model in generating memes for real-world examples (Tweets) through human evaluation. 

%Transformer variants
\begin{figure*}
    \centering
    \includegraphics[width=\linewidth]{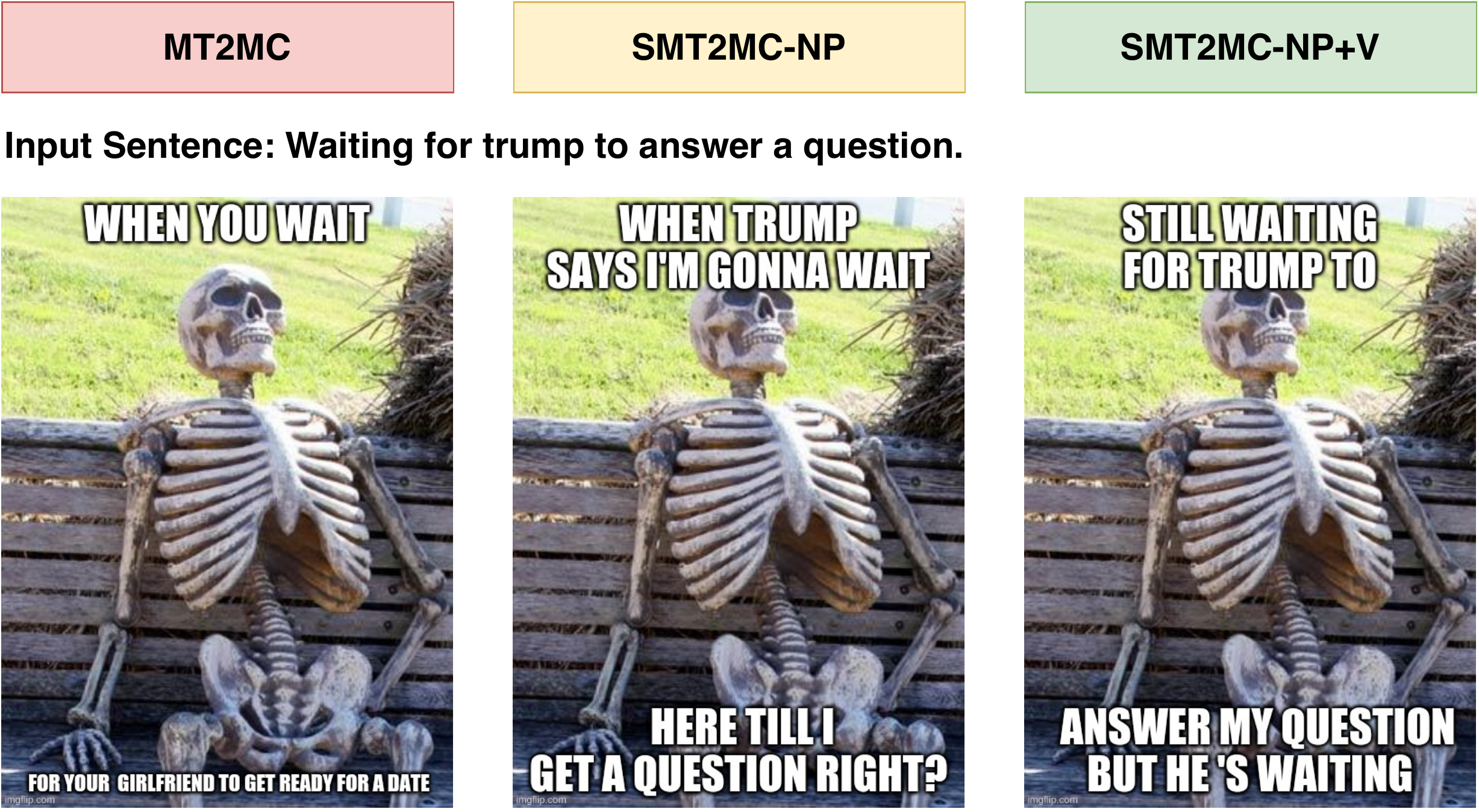}
    \caption{Memes generated by the caption generator variants for the given input sentence.}
     \label{fig:template_to_caption_result}
\end{figure*}

\subsection{Meme Template Selection Module}

We fine tune the pre-trained language representation models (BERT$_{base}$ \citep{devlin-etal-2019-bert}, XLNet$_{base}$ \citep{yang2019xlnet} and Roberta$_{base}$ \citep{liu2019roberta}) using a single layer linear neural network with 768 units on the meme template selection task using the meme caption dataset. Performance of the meme template selection module on the meme caption test data using variants of pre-trained language representation models is reported in Table ~\ref{tab:class_score}.
    
\begin{table}[h]
    \small
    \centering
    \begin{tabu}{cccc}
        \tabucline[1pt]-
        Model & Accuracy $\uparrow$ & F1 $\uparrow$ & Loss $\downarrow$\\
        \tabucline[1pt]-
        BERT$_{base}$&$68.18$ & $68.71$ & $1.15$\\ 
        \hline
        XLNet$_{base}$&$68.74$ & $69.06$ & $1.17$\\ 
        \hline
        Roberta$_{base}$&\textbf{69.31} & \textbf{69.87} & \textbf{1.10}\\ 
        \tabucline[1pt]-
    \end{tabu}
    \caption{Meme template selection module performance on the meme caption test dataset using the fine tuned language representation models. \textbf{Bold font} highlights the best scores obtained. For accuracy and F1, higher the score is better. For loss, lower the score is better.}
    \label{tab:class_score}
\end{table}

We adopt the best-performing model with a fine tuned Roberta$_{base}$ model as the meme template selection module in the meme generation pipeline.
   
\subsection{Meme Caption Generator}
For meme caption generation, we experiment with two different variants. The first variant - Meme Template to Meme Caption (MT2MC), inputs the selected meme template and generates a meme caption. The second variant - Sentence and Meme Template to Meme Caption (SMT2MC), inputs the input sentence along with the selected meme template and generates a meme caption. We use two variants as a part of ablation study to demonstrate the usage of the input sentence features enabling our model to generate memes relevant to the input sentence.

We also experiment with two variants of SMT2MC. The first variant uses only the noun phrases from the input sentence while the second variant uses the verbs along with the noun phrases. We experiment only using the noun phrases in order to study to what extent the addition of verbs directs the context of the generated meme towards the context of the input sentence. SMT2MC and MT2MC architectures follow the same denotations as of \citet{vaswani2017attention}. We report the hyper-parameters used in Table ~\ref{tbl:decoder}.

\begin{table}[h!]
    \small
    \centering
     \begin{tabu}{ccccc} 
        \tabucline[1pt]-
        Architecture & N & $d_{model}$ & $d_{ff}$ & h \\
        \tabucline[1pt]-
        MT2MC &8 & 768 & 2048 & 12\\
        \hline
        SMT2MC &6 & 512 & 2048 & 8 \\
        \tabucline[1pt]-
    \end{tabu}
    \caption{Hyper-parameters used in the caption generation models.}
    \label{tbl:decoder}
\end{table}

We use residual dropout ($P_{drop}$) \citep{srivastava2014dropout} for regularization and Adam optimizer \citep{kingma2014adam} with $\beta_{1} = 0.9, \beta_{2} = 0.98$ and $\epsilon=1e^{-9}$ and a scheduler using cosine annealing with warm restarts \citep{loshchilov2016sgdr}. During inference we generate meme captions using Beam search with a beam of size 6 and length penalty $\alpha = 0.7$. We stop the caption generation when a special end token or the maximum length of 32 tokens is reached. 

A sample of memes generated by the caption generator variants are presented in Figure ~\ref{fig:template_to_caption_result}. It can be seen that MT2MC generates a random caption for the given meme template which is irrelevant to the input sentence while the meme captions generated by the variants of SMT2MC are contextually relevant to the input sentence. Among the variants of SMT2MC, it can seen that the caption generated using noun phrases $\&$ verbs as inputs better represent the input sentence.

%BLEU score
\begin{table*}[h]
    \small
    \centering
    \begin{tabu}{ccccc}
        \tabucline[1pt]-
        Variant & BLEU-1 & BLEU-2 & BLEU-3  & BLEU-4\\
        \tabucline[1pt]-
        MT2MC &  $13.93$& $6.85.86$ & $4.45$ & $2.72$  \\ 
        \hline
        SMT2MC-NP & $38.71$& $24.67$ & $14.56$ & $8.89$  \\ 
        \hline
        SMT2MC-NP+V & \textbf{45.67} & \textbf{27.56} & \textbf{17.14} & \textbf{11.12}  \\ 
        \tabucline[1pt]-
    \end{tabu}
    \caption{BLEU scores for the caption generator variants. \textbf{Bold font} highlights the best scores obtained. NP - Noun Phrase and V - Verb.}
    \label{tab:automated_metrics}
\end{table*}

\begin{figure*}[h]
    \centering
    \subfloat[Coherence score distribution]{{\includegraphics[width=0.3\textwidth]{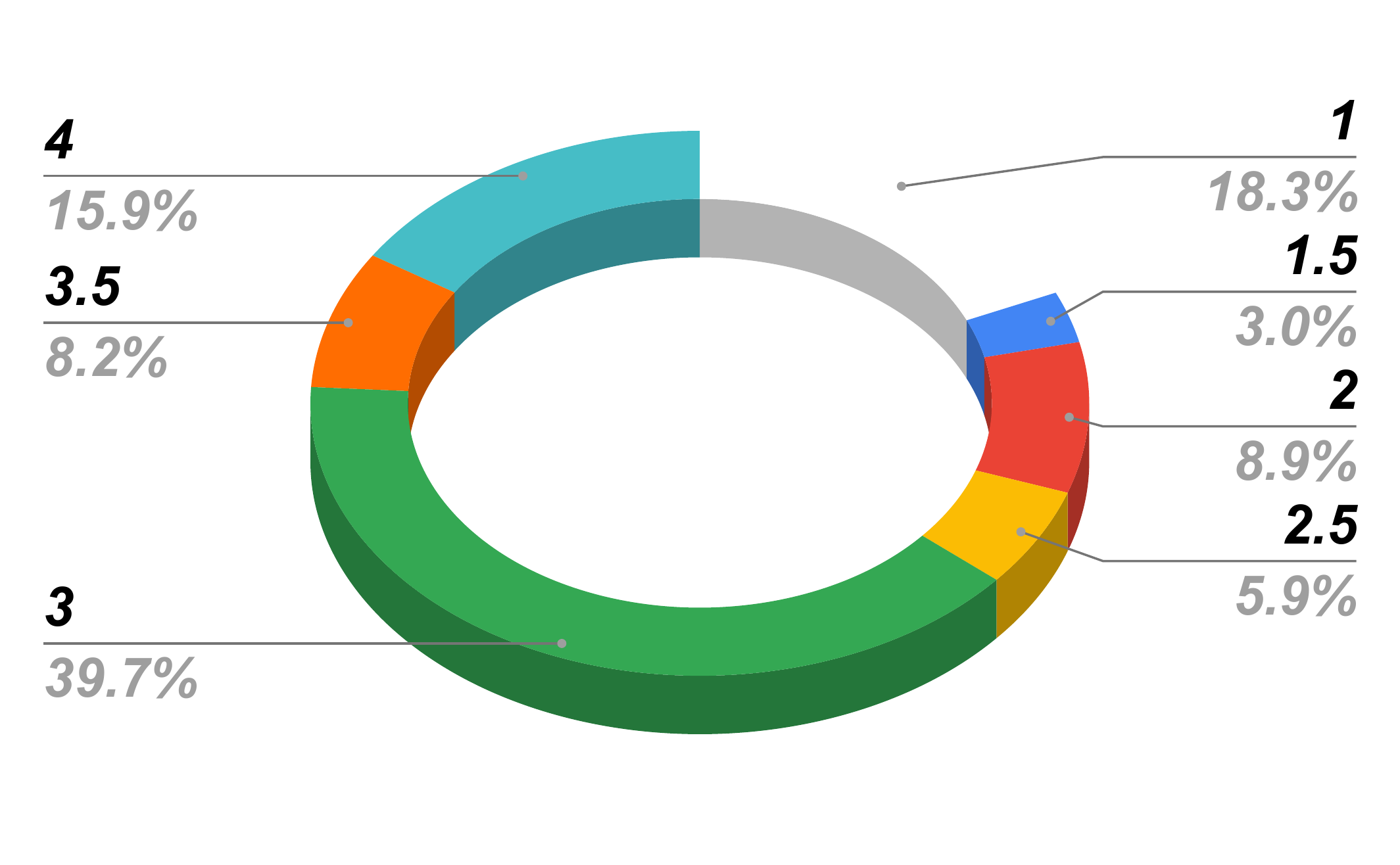} }}
    \subfloat[Relevance score distribution]{{\includegraphics[width=0.3\textwidth]{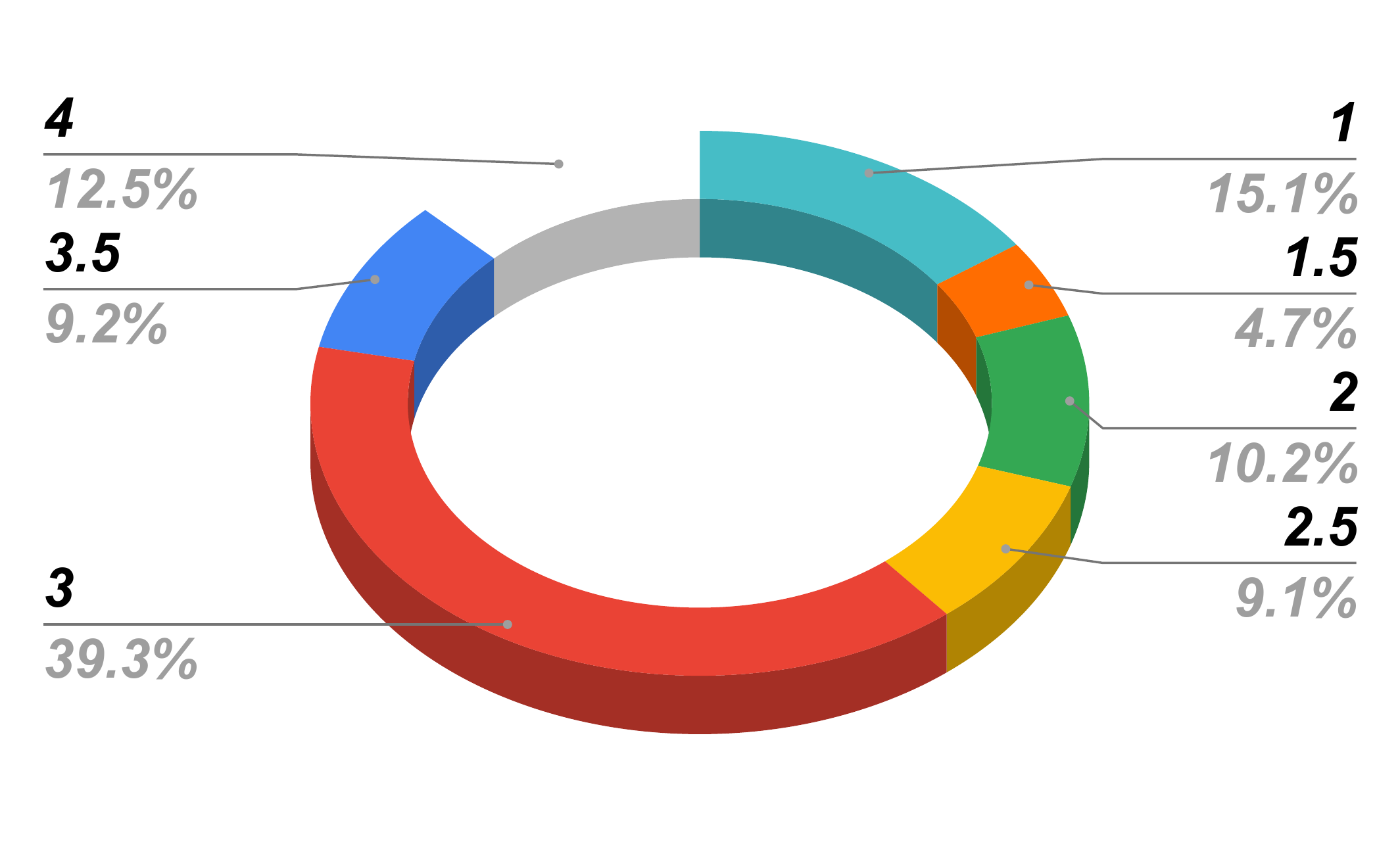} }}
    \subfloat[User Likes score distribution]{{\includegraphics[width=0.3\textwidth]{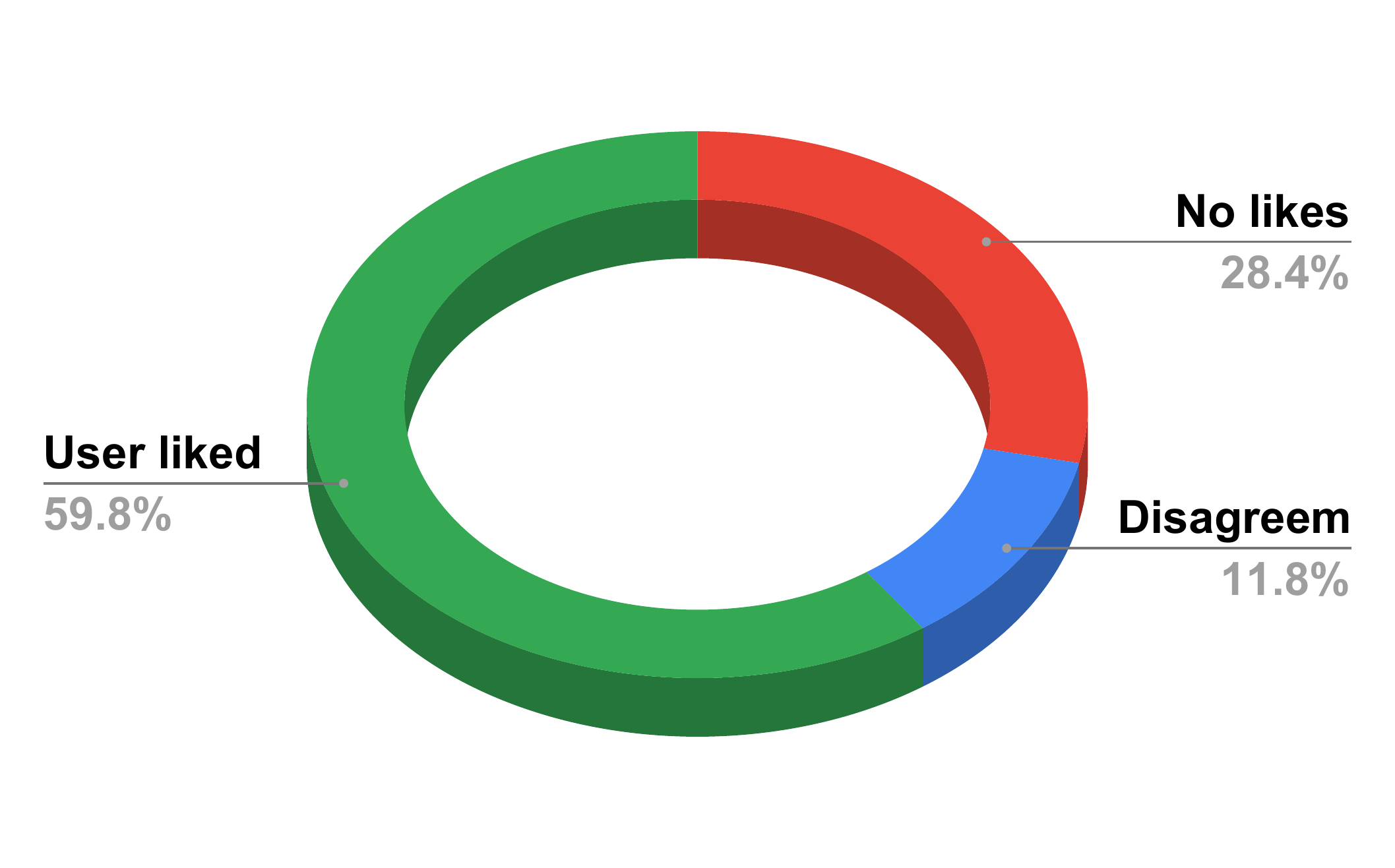} }}
    \caption{Human evaluation scores.}
    \label{fig:score_distribution}
\end{figure*}

\subsection{Evaluation Metrics}

We use BLEU score \citep{papineni2002bleu} to evaluate the quality of the generated captions. The perspective of good quality of a meme is subjective and varies among people. To the best of our knowledge, there are no known automatic evaluation metrics to evaluate the quality of a meme. A fairly reliable technique is to perform human evaluation by a set of raters to evaluate the quality of a meme on a subjective score.

\begin{figure*}[t]
  \centering
  \includegraphics[width=\textwidth]{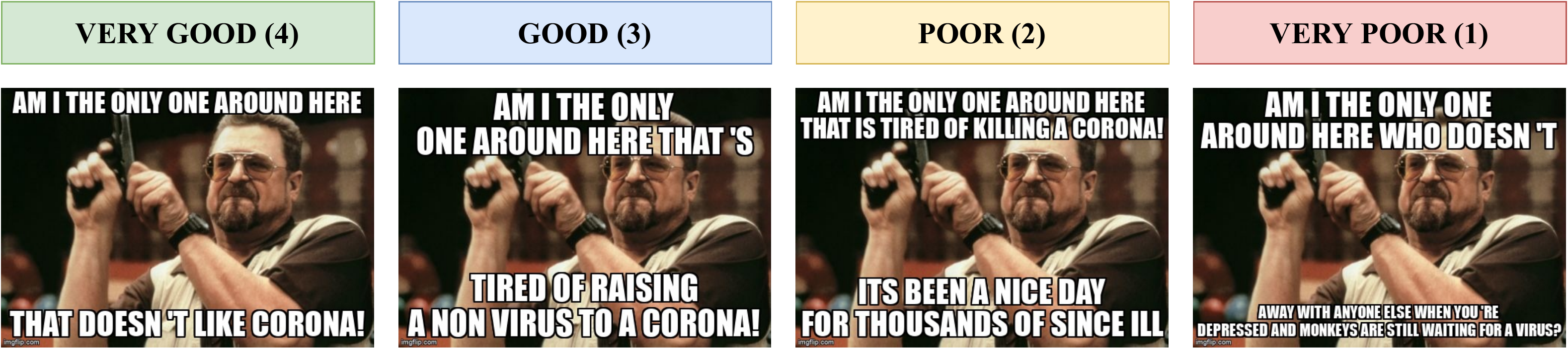}
   \caption{Qualitative comparison of memes grouped by Coherence score.}
   \label{fig:qualitative_figure_coherence}
\end{figure*}

\begin{figure*}[h]
  \centering
  \includegraphics[width=\textwidth]{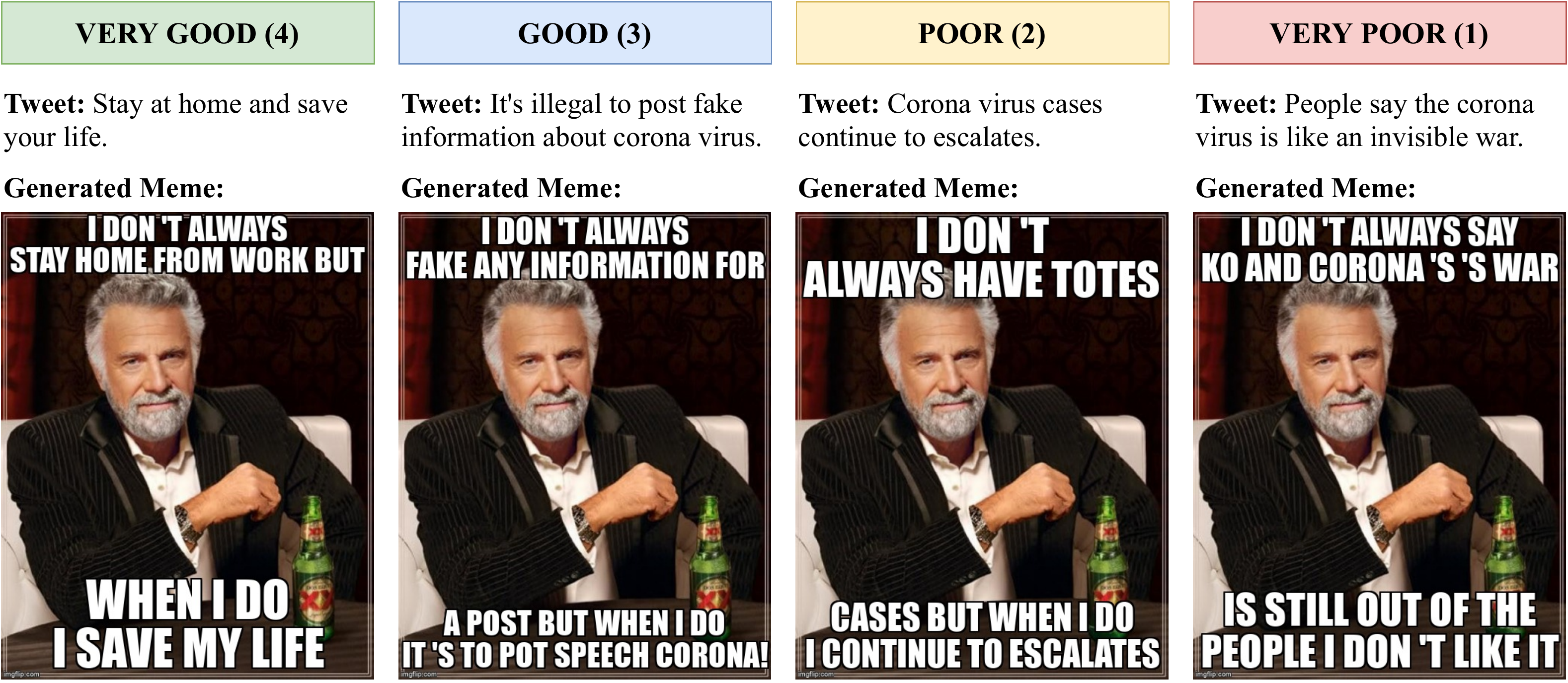}
  \caption{Qualitative comparison of memes grouped by Relevance score.}
  \label{fig:qualitative_figure_relevance}
\end{figure*}

In machine translation, adequacy and fluency \citep{snover2009fluency} are used to subjectively rate the correctness and fluency of a translation. Inspired from adequacy and fluency, we define 2 metrics - \textbf{Coherence and Relevance} to evaluate the generated memes, described as follows:
 
\begin{itemize}[noitemsep]
    \item  \textbf{Coherence:} Can you understand the message conveyed through the meme (Image + text)?
    \item \textbf{Relevance:} Is the meme contextually relevant to the text?
\end{itemize}
        
Coherence score captures the quality (fluency) of the generated meme and the Relevance score captures how well the generated meme represents the input sentence (correctness). We also ask the raters if they like the meme to evaluate if the generated memes are good. The Relevance and Coherence metrics are scored on a range of 1 - 4. \textbf{User Likes} score represents the percentage of total raters who liked the meme. To score these metrics, we set up an Amazon Mechanical Turk (AMT) experiment.

\subsection{Evaluation Results}

\subsubsection{Caption Generation Results}

The scores for the caption generator variants are reported in Table \ref{tab:automated_metrics}. We see that the SMT2MC variants produce meme captions textually similar to that of the input sentence. Among the SMT2MC variants, the variant which inputs verbs along with noun phrases has better score, and using verbs with noun phrases enables the caption generator to generate relatively relevant captions to that of the input sentence when compared to the variant which inputs only the noun phrases. We use the best performing SMT2MC-NP+V to generate memes for Twitter data.

\subsubsection{Human Evaluation Task Setup}

We choose Amazon Mechanical Turk (AMT) for the evaluation of the generated memes due to its easy to use platform and the ready availability of a big worker pool with required skills. An example for AMT questionnaire is in presented in Appendix C. Each sample was rated by 2 raters and in case of disagreement among the raters, we consider their average score as the final score.

In our AMT evaluation setup, we design a two-stage process to evaluate the meme. We first display the meme image and ask the workers to score the Coherence metric, only based on their understanding of the meme. Later we display the tweet and ask them to understand the text, and then ask them to score the Relevance metric based on their comprehension of the tweet and the meme. Our expectation for the AMT workers is that they are capable of visually understanding an image, capable of semantically and contextually understanding a sentence and possess the reasoning ability to compare context from different information sources. We assume an adult human being is well qualified to meet our expectations. 

%results figure
\begin{figure*}[h]
    \centering
    \includegraphics[width=0.9\linewidth]{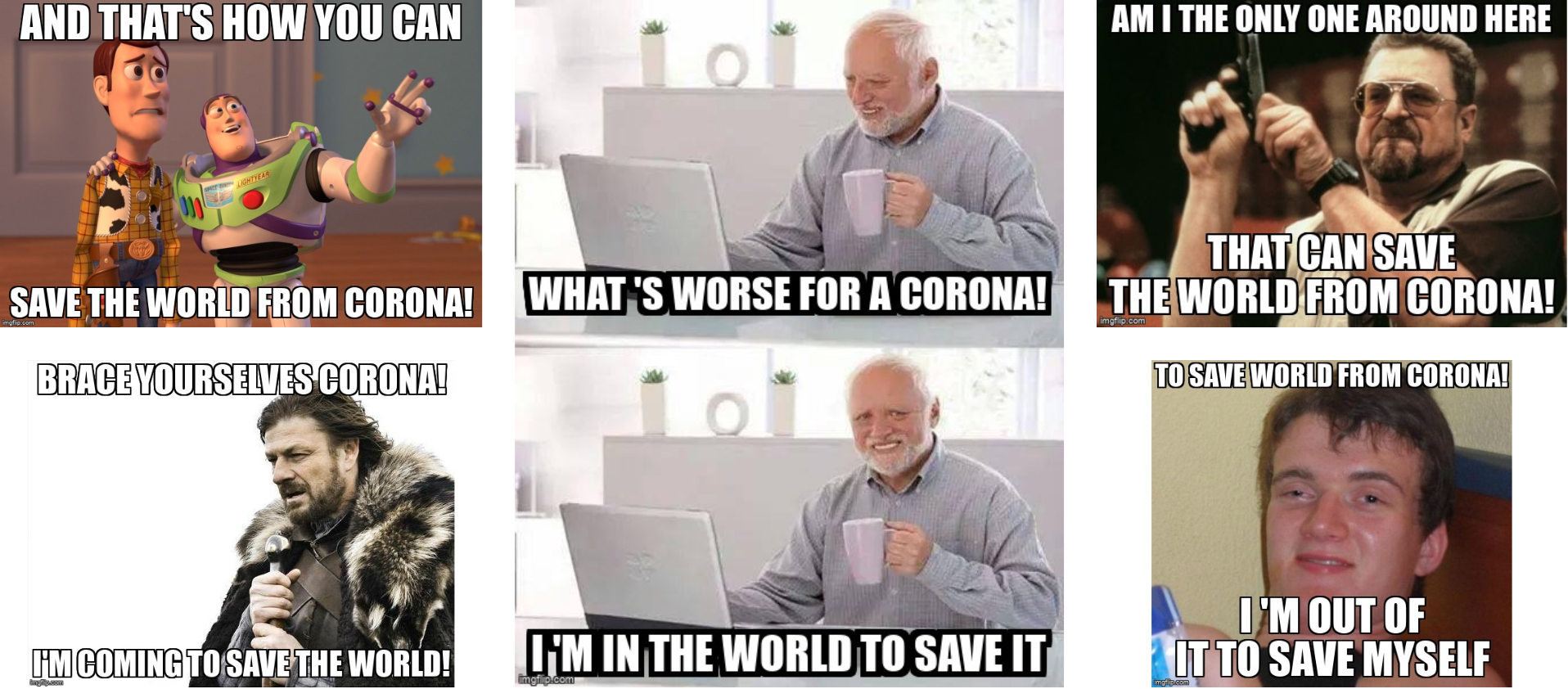}
    \caption{Memes generated by the SMT2MC-NP+V variant for the input tweet -``Please save the world from Corona" by conditioning caption generation using different meme templates.}
     \label{fig:ctrl}
\end{figure*}

\subsubsection{Human Evaluation Results}
 The performance of the SMT2MC-NP+V model on the human evaluation metrics is reported in Table \ref{tbl:human_eval} and the score distribution across the evaluation metrics is presented in the Figure \ref{fig:score_distribution}. A qualitative comparison of memes grouped by rater scores is presented in Figures ~\ref{fig:qualitative_figure_coherence} and ~\ref{fig:qualitative_figure_relevance}.

\begin{table}[h]
    \small
    \centering
    \begin{tabu}{cc}
        \tabucline[1pt]-
        Metric & Score\\
        \tabucline[1pt]-
        Coherence & $2.66$\\
        \hline
        Relevance & $2.65$\\
        \hline
        User Likes & $0.65$\\
        \tabucline[1pt]-
    \end{tabu}
    \caption{Human evaluation scores on twitter data. Relevance and Coherence metrics are scored out of 4. User Likes score represents the percentage of total raters who liked the meme.}
    \label{tbl:human_eval}
\end{table}

Before interpreting the scores, we review the image meme generation task. It requires the ability to semantically and contextually understand the input sentence along with the contextual knowledge of the image memes. Even with the understanding of the input sentence and meme images, one has to posses a good fluency in natural language to generate a meme caption that is compatible with the meme image. The generated meme should also be relevant to the input sentence. We analyze the performance of our model by assuming that a human generated meme would get perfect score across all the metrics in generating a good quality meme for a tweet.

Observing the score distribution from Figure \ref{fig:score_distribution}, we infer that more than $60\%$ of the generated memes are coherent and relevant to the input tweets. From Table \ref{tbl:human_eval}, we see that $65\%$ of the raters like the meme shown to them and the like percentage correlate with the coherence and relevance scores. We infer that the raters have liked the meme if they understood the information conveyed through the meme and if the meme is relevant to the input tweet. Quantitatively, our model is capable of generating coherent memes with $66.5\%$ confidence and relevant memes with $66.25\%$ confidence. Our model performs with good confidence on the challenging image meme generation task using only the language features of the image meme during training.

\section{Inter Rater Reliability}

We use Cohen's Kappa $(\kappa)$ to measure the reliability among the raters. Cohen's Kappa is defined as
\begin{equation}
    \kappa = \frac{p_{o} - P_{e}}{N-P_{e}}
\end{equation}

where $p_{o}$ is the relative observed agreement and $p_{e}$ is the hypothetical probability of chance agreement among the raters, and N is the number of samples. The Inter Rater Reliability (IRR) score among the users on different metrics is reported in Table ~\ref{tbl:irr}.

\begin{table}[h]
    \small
    \centering
    \begin{tabu}{cc}
        \tabucline[1pt]-
        Metric & Agreement Score\\
        \tabucline[1pt]-
        Coherence & $71.68$\\
        \hline
        Relevance & $61.39$\\
        \hline
        User Likes & $73.85$\\
        \tabucline[1pt]-
    \end{tabu}
    \caption{Inter Rater Reliability scores [$\%$] on the human evaluation metrics.}
    \label{tbl:irr}
\end{table}

The raters have higher than $60\%$ agreement on all the metrics which establishes a good consistency among the raters for evaluating the quality of the generated image memes.

\section{Controlling Meme Generation}

Corrupting the input data during training enables our model to learn from the meme caption dataset and scale our model for any input sentence during inference as shown in Figure  \ref{fig:qualitative_figure_relevance}. In an ideal scenario, the user might want to select the meme template. During the experiments, we observed that for a given sentence, information abstraction during training has enabled our model to create a meme caption conditioned on any given meme template. We experiment further on this by forcing the caption generator to generate captions for a input sentence conditioned on different meme templates. The generated memes are presented in Figure \ref{fig:ctrl} and our model is capable of generating a meme for an input sentence conditioned on a selected or a given meme template.

\section{Conclusion and Future Work}

We have presented memeBot, an end to end architecture that can automatically generate a meme for a given sentence. memeBot is composed of two components, a module to select a meme template and an encoder-decoder to generate a meme caption. The model is trained on a meme caption dataset to maximize the likelihood of selecting a template given a caption and to maximize the likelihood of generating a meme caption given the input sentence and the meme template. Automatic evaluation on meme caption test data and human evaluation scores on Twitter data show promising performance in generating an image for sentences in online social interaction.

The concept of quality of a meme highly varies among people and is hard to evaluate using a set of pre-defined metrics. In real-world scenarios, if an individual likes a meme, he or she shares it with others. If a group of individuals like the same meme then the meme can become viral or trending. Future work includes evaluating a meme by introducing it in a social media stream and rate the meme based on its transmission among the people. The meme transmission rate and the group of people it transmits across can be used as reinforcement to generate more creative and better quality meme. \newline

\noindent\textbf{Acknowledgment:} The Office of Naval Research (award \#N00014-18-1-2761) and the National Science Foundation under the Robust Intelligence Program (\#1750082) are gratefully acknowledged.

\bibliography{emnlp2020}

\begin{thebibliography}{27}
\expandafter\ifx\csname natexlab\endcsname\relax\def\natexlab#1{#1}\fi

\bibitem[{Bahdanau et~al.(2014)Bahdanau, Cho, and Bengio}]{nmt}
Dzmitry Bahdanau, Kyunghyun Cho, and Yoshua Bengio. 2014.
\newblock Neural machine translation by jointly learning to align and
  translate.
\newblock \emph{arXiv preprint arXiv:1409.0473}.

\bibitem[{Davison(2012)}]{davison2012language}
Patrick Davison. 2012.
\newblock The language of internet memes.
\newblock \emph{The social media reader}, pages 120--134.

\bibitem[{Devlin et~al.(2019)Devlin, Chang, Lee, and
  Toutanova}]{devlin-etal-2019-bert}
Jacob Devlin, Ming-Wei Chang, Kenton Lee, and Kristina Toutanova. 2019.
\newblock \href {https://doi.org/10.18653/v1/N19-1423} {{BERT}: Pre-training of
  deep bidirectional transformers for language understanding}.
\newblock In \emph{Proceedings of the 2019 Conference of the North {A}merican
  Chapter of the Association for Computational Linguistics: Human Language
  Technologies, Volume 1 (Long and Short Papers)}, pages 4171--4186,
  Minneapolis, Minnesota. Association for Computational Linguistics.

\bibitem[{Fang et~al.(2020)Fang, Gokhale, Banerjee, Baral, and
  Yang}]{fang2020video2commonsense}
Zhiyuan Fang, Tejas Gokhale, Pratyay Banerjee, Chitta Baral, and Yezhou Yang.
  2020.
\newblock Video2commonsense: Generating commonsense descriptions to enrich
  video captioning.
\newblock \emph{arXiv preprint arXiv:2003.05162}.

\bibitem[{Honnibal and Montani(2017)}]{spacy2}
Matthew Honnibal and Ines Montani. 2017.
\newblock {spaCy 2}: Natural language understanding with {B}loom embeddings,
  convolutional neural networks and incremental parsing.
\newblock To appear.

\bibitem[{Hu et~al.(2017)Hu, Yang, Liang, Salakhutdinov, and
  Xing}]{hu2017toward}
Zhiting Hu, Zichao Yang, Xiaodan Liang, Ruslan Salakhutdinov, and Eric~P Xing.
  2017.
\newblock Toward controlled generation of text.
\newblock In \emph{Proceedings of the 34th International Conference on Machine
  Learning-Volume 70}, pages 1587--1596. JMLR. org.

\bibitem[{Huang et~al.(2018)Huang, Zaiane, Trabelsi, and
  Dziri}]{huang2018automatic}
Chenyang Huang, Osmar Zaiane, Amine Trabelsi, and Nouha Dziri. 2018.
\newblock Automatic dialogue generation with expressed emotions.
\newblock In \emph{Proceedings of the 2018 Conference of the North American
  Chapter of the Association for Computational Linguistics: Human Language
  Technologies, Volume 2 (Short Papers)}, pages 49--54.

\bibitem[{Hutto and Gilbert(2014)}]{hutto2014vader}
Clayton~J Hutto and Eric Gilbert. 2014.
\newblock Vader: A parsimonious rule-based model for sentiment analysis of
  social media text.
\newblock In \emph{Eighth international AAAI conference on weblogs and social
  media}.

\bibitem[{Karpathy and Fei-Fei(2015)}]{imgdesc}
Andrej Karpathy and Li~Fei-Fei. 2015.
\newblock Deep visual-semantic alignments for generating image descriptions.
\newblock In \emph{Proceedings of the IEEE conference on computer vision and
  pattern recognition}, pages 3128--3137.

\bibitem[{Kingma and Ba(2014)}]{kingma2014adam}
Diederik~P Kingma and Jimmy Ba. 2014.
\newblock Adam: A method for stochastic optimization.
\newblock \emph{arXiv preprint arXiv:1412.6980}.

\bibitem[{Liu et~al.(2019)Liu, Ott, Goyal, Du, Joshi, Chen, Levy, Lewis,
  Zettlemoyer, and Stoyanov}]{liu2019roberta}
Yinhan Liu, Myle Ott, Naman Goyal, Jingfei Du, Mandar Joshi, Danqi Chen, Omer
  Levy, Mike Lewis, Luke Zettlemoyer, and Veselin Stoyanov. 2019.
\newblock Roberta: A robustly optimized bert pretraining approach.
\newblock \emph{arXiv preprint arXiv:1907.11692}.

\bibitem[{Loshchilov and Hutter(2016)}]{loshchilov2016sgdr}
Ilya Loshchilov and Frank Hutter. 2016.
\newblock Sgdr: Stochastic gradient descent with warm restarts.
\newblock \emph{arXiv preprint arXiv:1608.03983}.

\bibitem[{Luong et~al.(2015)Luong, Pham, and
  Manning}]{luong-etal-2015-effective}
Thang Luong, Hieu Pham, and Christopher~D. Manning. 2015.
\newblock \href {https://doi.org/10.18653/v1/D15-1166} {Effective approaches to
  attention-based neural machine translation}.
\newblock In \emph{Proceedings of the 2015 Conference on Empirical Methods in
  Natural Language Processing}, pages 1412--1421, Lisbon, Portugal. Association
  for Computational Linguistics.

\bibitem[{Miao et~al.(2019)Miao, Zhou, Mou, Yan, and Li}]{miao2019cgmh}
Ning Miao, Hao Zhou, Lili Mou, Rui Yan, and Lei Li. 2019.
\newblock Cgmh: Constrained sentence generation by metropolis-hastings
  sampling.
\newblock In \emph{Proceedings of the AAAI Conference on Artificial
  Intelligence}, volume~33, pages 6834--6842.

\bibitem[{Papineni et~al.(2002)Papineni, Roukos, Ward, and
  Zhu}]{papineni2002bleu}
Kishore Papineni, Salim Roukos, Todd Ward, and Wei-Jing Zhu. 2002.
\newblock Bleu: a method for automatic evaluation of machine translation.
\newblock In \emph{Proceedings of the 40th annual meeting on association for
  computational linguistics}, pages 311--318. Association for Computational
  Linguistics.

\bibitem[{Peirson et~al.(2018)Peirson, Abel, and Tolunay}]{peirson2018dank}
V~Peirson, L~Abel, and E~Meltem Tolunay. 2018.
\newblock Dank learning: Generating memnlp papes using deep neural networks.
\newblock \emph{arXiv preprint arXiv:1806.04510}.

\bibitem[{Snover et~al.(2009)Snover, Madnani, Dorr, and
  Schwartz}]{snover2009fluency}
Matthew Snover, Nitin Madnani, Bonnie~J Dorr, and Richard Schwartz. 2009.
\newblock Fluency, adequacy, or hter?: exploring different human judgments with
  a tunable mt metric.
\newblock In \emph{Proceedings of the Fourth Workshop on Statistical Machine
  Translation}, pages 259--268. Association for Computational Linguistics.

\bibitem[{Socher et~al.(2013)Socher, Perelygin, Wu, Chuang, Manning, Ng, and
  Potts}]{socher2013recursive}
Richard Socher, Alex Perelygin, Jean Wu, Jason Chuang, Christopher~D Manning,
  Andrew Ng, and Christopher Potts. 2013.
\newblock Recursive deep models for semantic compositionality over a sentiment
  treebank.
\newblock In \emph{Proceedings of the 2013 conference on empirical methods in
  natural language processing}, pages 1631--1642.

\bibitem[{Srivastava et~al.(2014)Srivastava, Hinton, Krizhevsky, Sutskever, and
  Salakhutdinov}]{srivastava2014dropout}
Nitish Srivastava, Geoffrey Hinton, Alex Krizhevsky, Ilya Sutskever, and Ruslan
  Salakhutdinov. 2014.
\newblock Dropout: a simple way to prevent neural networks from overfitting.
\newblock \emph{The Journal of Machine Learning Research}, 15(1):1929--1958.

\bibitem[{Su et~al.(2018)Su, Xu, Qiu, and Huang}]{su2018incorporating}
Jinyue Su, Jiacheng Xu, Xipeng Qiu, and Xuanjing Huang. 2018.
\newblock Incorporating discriminator in sentence generation: a gibbs sampling
  method.
\newblock In \emph{Thirty-Second AAAI Conference on Artificial Intelligence}.

\bibitem[{Sutskever et~al.(2014)Sutskever, Vinyals, and
  Le}]{sutskever2014sequence}
Ilya Sutskever, Oriol Vinyals, and Quoc~V Le. 2014.
\newblock Sequence to sequence learning with neural networks.
\newblock In \emph{Advances in neural information processing systems}, pages
  3104--3112.

\bibitem[{Vaswani et~al.(2017)Vaswani, Shazeer, Parmar, Uszkoreit, Jones,
  Gomez, Kaiser, and Polosukhin}]{vaswani2017attention}
Ashish Vaswani, Noam Shazeer, Niki Parmar, Jakob Uszkoreit, Llion Jones,
  Aidan~N Gomez, {\L}ukasz Kaiser, and Illia Polosukhin. 2017.
\newblock Attention is all you need.
\newblock In \emph{Advances in neural information processing systems}, pages
  5998--6008.

\bibitem[{Vincent et~al.(2008)Vincent, Larochelle, Bengio, and
  Manzagol}]{vincent2008extracting}
Pascal Vincent, Hugo Larochelle, Yoshua Bengio, and Pierre-Antoine Manzagol.
  2008.
\newblock Extracting and composing robust features with denoising autoencoders.
\newblock In \emph{Proceedings of the 25th international conference on Machine
  learning}, pages 1096--1103. ACM.

\bibitem[{Vinyals et~al.(2015)Vinyals, Toshev, Bengio, and Erhan}]{showtell}
Oriol Vinyals, Alexander Toshev, Samy Bengio, and Dumitru Erhan. 2015.
\newblock Show and tell: A neural image caption generator.
\newblock In \emph{Proceedings of the IEEE conference on computer vision and
  pattern recognition}, pages 3156--3164.

\bibitem[{Wang and Wen(2015)}]{wang2015can}
William~Yang Wang and Miaomiao Wen. 2015.
\newblock I can has cheezburger? a nonparanormal approach to combining textual
  and visual information for predicting and generating popular meme
  descriptions.
\newblock In \emph{Proceedings of the 2015 Conference of the North American
  Chapter of the Association for Computational Linguistics: Human Language
  Technologies}, pages 355--365.

\bibitem[{Xu et~al.(2015)Xu, Ba, Kiros, Cho, Courville, Salakhudinov, Zemel,
  and Bengio}]{showattendtell}
Kelvin Xu, Jimmy Ba, Ryan Kiros, Kyunghyun Cho, Aaron Courville, Ruslan
  Salakhudinov, Rich Zemel, and Yoshua Bengio. 2015.
\newblock Show, attend and tell: Neural image caption generation with visual
  attention.
\newblock In \emph{International conference on machine learning}, pages
  2048--2057.

\bibitem[{Yang et~al.(2019)Yang, Dai, Yang, Carbonell, Salakhutdinov, and
  Le}]{yang2019xlnet}
Zhilin Yang, Zihang Dai, Yiming Yang, Jaime Carbonell, Ruslan Salakhutdinov,
  and Quoc~V Le. 2019.
\newblock Xlnet: Generalized autoregressive pretraining for language
  understanding.
\newblock \emph{arXiv preprint arXiv:1906.08237}.

\end{thebibliography}
\bibliographystyle{acl_natbib}

\appendix
\onecolumn

\section{Meme Templates}

\begin{table*}[h!]
    \begin{center}
         \begin{tabular}{|m{1.75cm} | c | m{1.75cm} | c | m{1.75cm} | c  | }
            \hline
            \centering
            \textbf{Template Name} 
            &
            \textbf{Template Image}
            &
            \centering
            \textbf{Template Name} 
            &
            \textbf{Template Image}
            &
            \centering
            \textbf{Template Name} 
            &
            \textbf{Template Image}
            \\ \hline
            
            \centering
            Bad Luck Brian
            &
            \begin{minipage}{.15\textwidth}
                \vspace{0.2cm}
                \begin{center}
                    \includegraphics[height= 1.5cm]{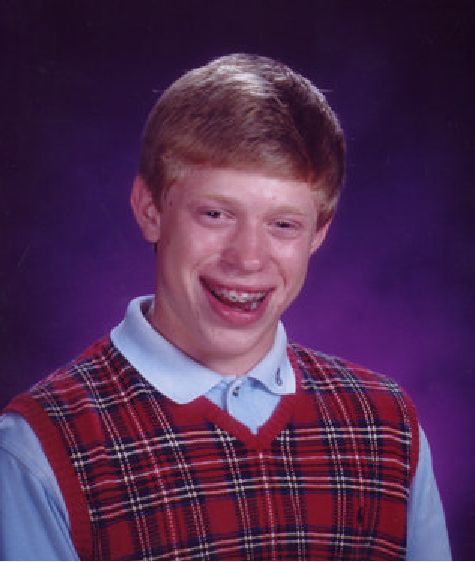}
                \end{center}
                \vspace{0.1cm}
            \end{minipage} 
            &
            \centering
            Leonardo Dicaprio Cheers
            &
            \begin{minipage}{.15\textwidth}
                \vspace{0.2cm}
                \begin{center}
                    \includegraphics[height= 1.5cm]{images/meme_templates/Leonardo-Dicaprio-Cheers.jpg}
                \end{center}
                \vspace{0.1cm}
            \end{minipage} 
            &
            \centering
            Success Kid
            &
            \begin{minipage}{.15\textwidth}
                \vspace{0.2cm}
                \begin{center}
                    \includegraphics[height= 1.5cm]{images/meme_templates/Success-Kid.jpg}
                \end{center}
                \vspace{0.1cm}
            \end{minipage} 
            \\ \hline
            
            \centering
            X X Everywhere
            &
            \begin{minipage}{.15\textwidth}
                \vspace{0.2cm}
                \begin{center}
                    \includegraphics[height= 1.5cm]{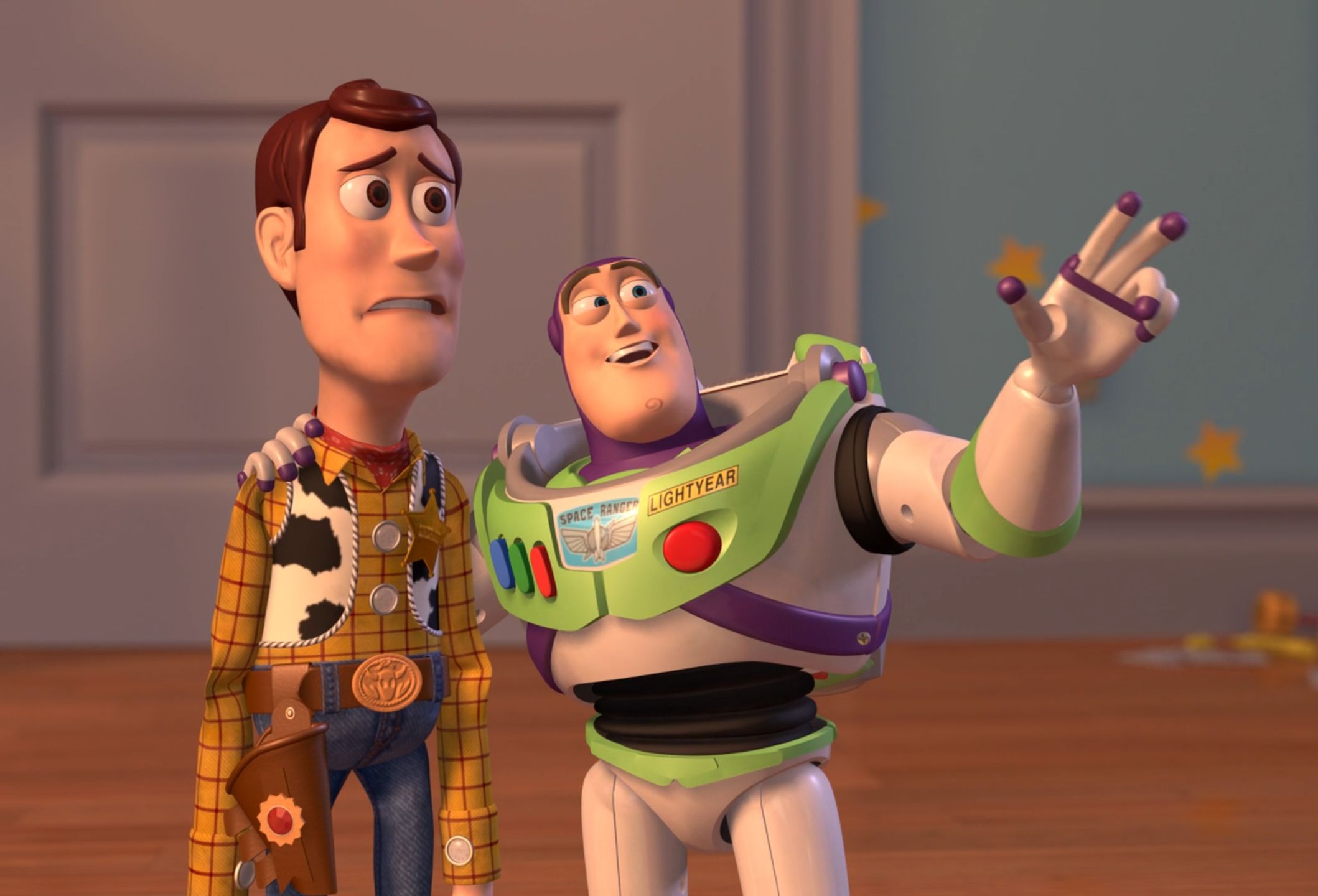}
                \end{center}
                \vspace{0.1cm}
            \end{minipage} 
            &
            \centering
            Ancient Aliens
            &
            \begin{minipage}{.15\textwidth}
                \vspace{0.2cm}
                \begin{center}
                    \includegraphics[height= 1.5cm]{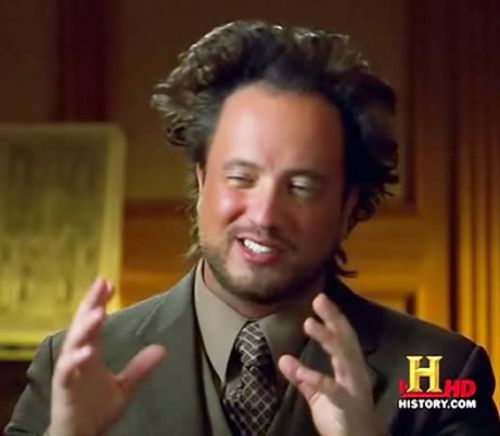}
                \end{center}
                \vspace{0.1cm}
            \end{minipage} 
            &
            \centering
            Disaster Girl
            &
            \begin{minipage}{.15\textwidth}
                \vspace{0.2cm}
                \begin{center}
                    \includegraphics[height= 1.5cm]{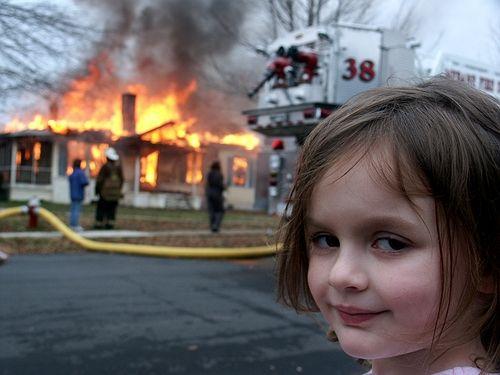}
                \end{center}
                \vspace{0.1cm}
            \end{minipage} 
            \\ \hline
            
            \centering
            One Does Not Simply
            &
            \begin{minipage}{.15\textwidth}
                \vspace{0.2cm}
                \begin{center}
                    \includegraphics[height= 1.5cm]{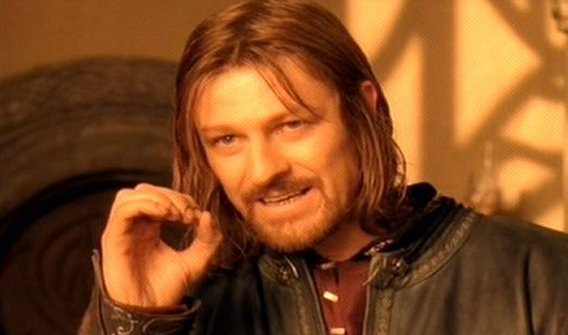}
                \end{center}
                \vspace{0.1cm}
            \end{minipage} 
            &
            \centering
            Third World Skeptical Kid
            &
            \begin{minipage}{.15\textwidth}
                \vspace{0.2cm}
                \begin{center}
                    \includegraphics[height= 1.5cm]{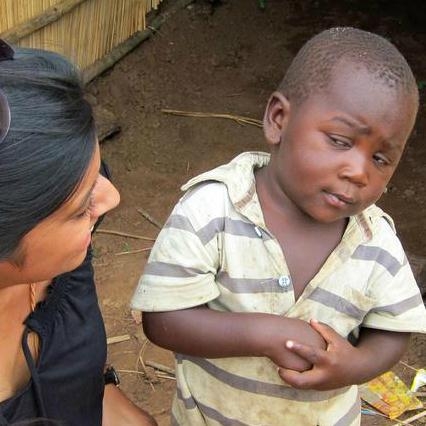}
                \end{center}
                \vspace{0.1cm}
            \end{minipage} 
            &
            \centering
            Futurama Fry
            &
            \begin{minipage}{.15\textwidth}
                \vspace{0.2cm}
                \begin{center}
                    \includegraphics[height= 1.5cm]{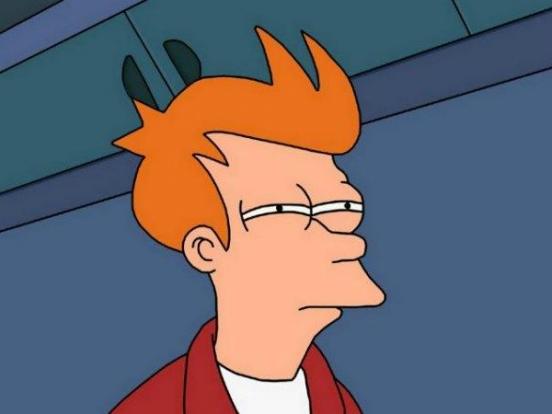}
                \end{center}
                \vspace{0.1cm}
            \end{minipage} 
            \\ \hline
            
            \centering
            10 Guy
            &
            \begin{minipage}{.15\textwidth}
                \vspace{0.2cm}
                \begin{center}
                    \includegraphics[height= 1.5cm]{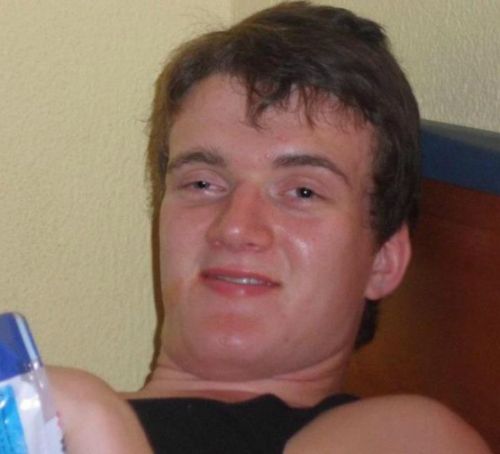}
                \end{center}
                \vspace{0.1cm}
            \end{minipage} 
            &
            \centering
            Am I The Only One Around Here
            &
            \begin{minipage}{.15\textwidth}
                \vspace{0.2cm}
                \begin{center}
                    \includegraphics[height= 1.5cm]{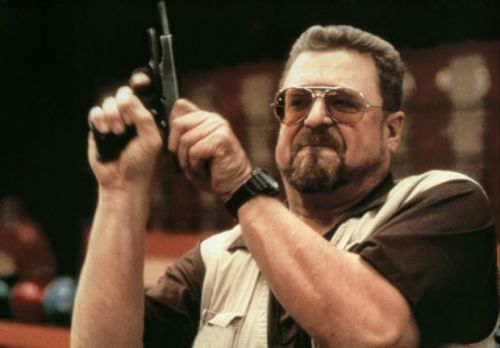}
                \end{center}
                \vspace{0.1cm}
            \end{minipage} 
            &
            \centering
            Captain Picard Facepalm
            &
            \begin{minipage}{.15\textwidth}
                \vspace{0.2cm}
                \begin{center}
                    \includegraphics[height= 1.5cm]{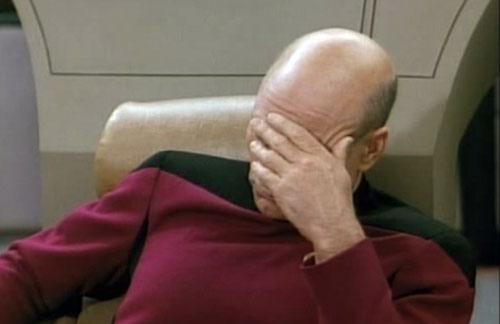}
                \end{center}
                \vspace{0.1cm}
            \end{minipage} 
            \\ \hline
            
            \centering
            Brace Yourselves X is Coming
            &
            \begin{minipage}{.15\textwidth}
                \vspace{0.2cm}
                \begin{center}
                    \includegraphics[height= 1.5cm]{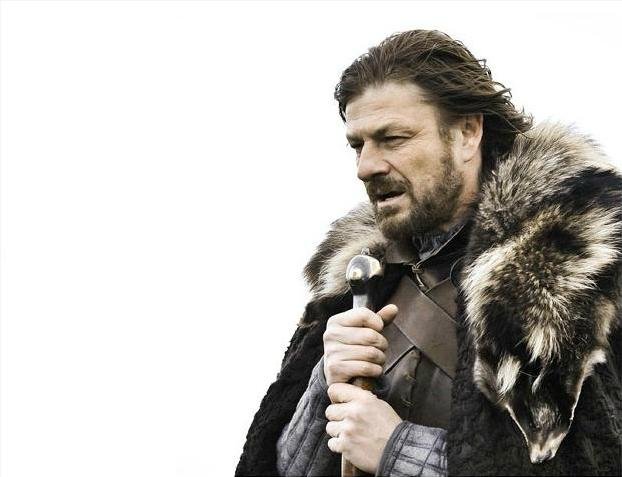}
                \end{center}
                \vspace{0.1cm}
            \end{minipage} 
            &
            \centering
            Creepy Condescending Wonka
            &
            \begin{minipage}{.15\textwidth}
                \vspace{0.2cm}
                \begin{center}
                    \includegraphics[height= 1.5cm]{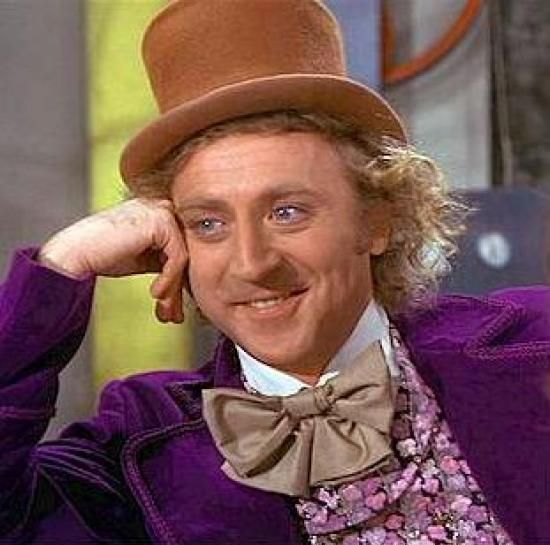}
                \end{center}
                \vspace{0.1cm}
            \end{minipage} 
            &
            \centering
            Matrix Morpheus
            &
            \begin{minipage}{.15\textwidth}
                \vspace{0.2cm}
                \begin{center}
                    \includegraphics[height= 1.5cm]{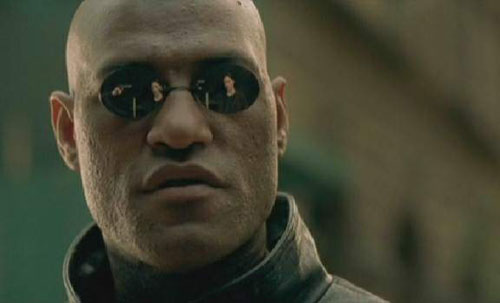}
                \end{center}
                \vspace{0.1cm}
            \end{minipage} 
            \\ \hline
            
            \centering
            Y U No
            &
            \begin{minipage}{.15\textwidth}
                \vspace{0.2cm}
                \begin{center}
                    \includegraphics[height= 1.5cm]{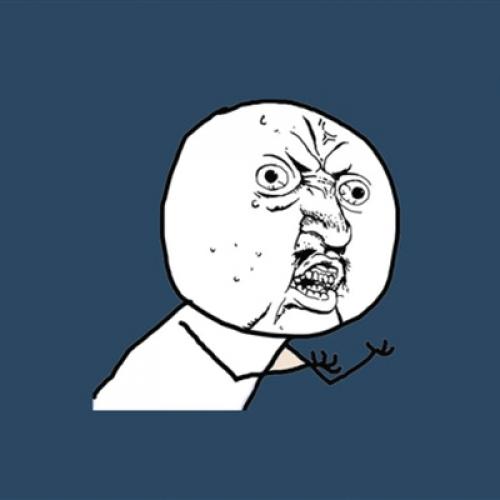}
                \end{center}
                \vspace{0.1cm}
            \end{minipage} 
            &
            \centering
            But Thats None Of My Business
            &
            \begin{minipage}{.15\textwidth}
                \vspace{0.2cm}
                \begin{center}
                    \includegraphics[height= 1.5cm]{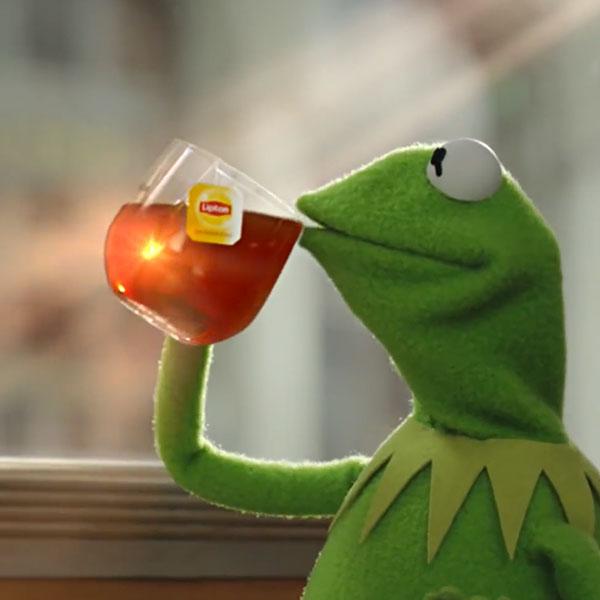}
                \end{center}
                \vspace{0.1cm}
            \end{minipage} 
            &
            \centering
            Roll Safe Think About It
            &
            \begin{minipage}{.15\textwidth}
                \vspace{0.2cm}
                \begin{center}
                    \includegraphics[height= 1.5cm]{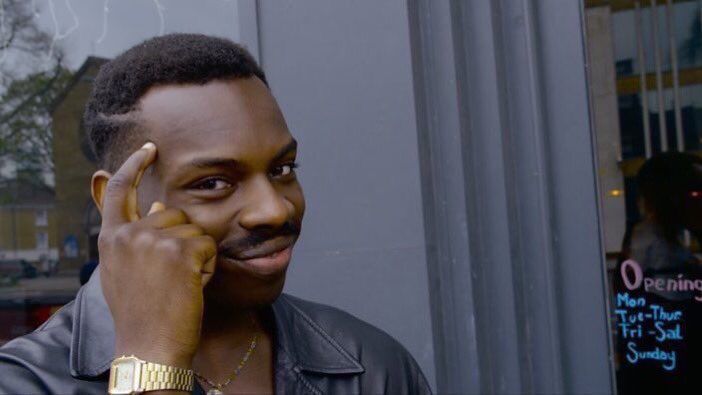}
                \end{center}
                \vspace{0.1cm}
            \end{minipage} 
            \\ \hline
            
            \centering
            Waiting Skeleton
            &
            \begin{minipage}{.15\textwidth}
                \vspace{0.2cm}
                \begin{center}
                    \includegraphics[height= 1.5cm]{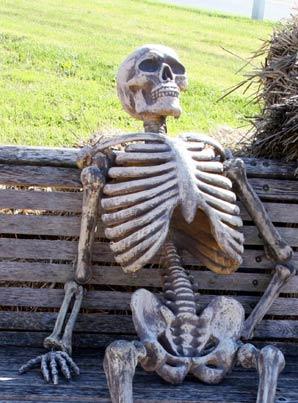}
                \end{center}
                \vspace{0.1cm}
            \end{minipage} 
            &
            \centering
            The Most Interesting Man In The World
            &
            \begin{minipage}{.15\textwidth}
                \vspace{0.2cm}
                \begin{center}
                    \includegraphics[height= 1.5cm]{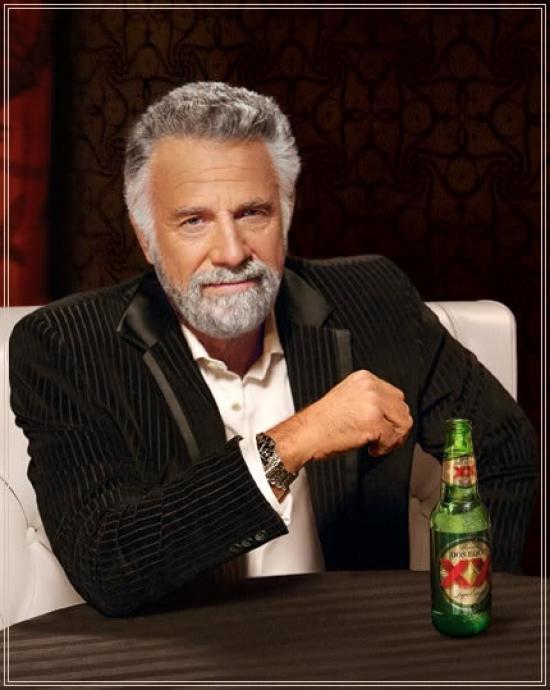}
                \end{center}
                \vspace{0.1cm}
            \end{minipage} 
            &
            \centering
            First World Problems
            &
            \begin{minipage}{.15\textwidth}
                \vspace{0.2cm}
                \begin{center}
                    \includegraphics[height= 1.5cm]{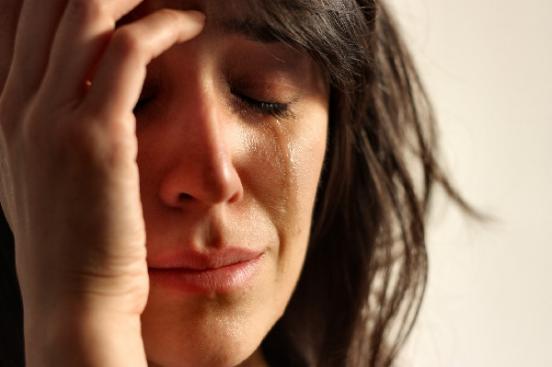}
                \end{center}
                \vspace{0.1cm}
            \end{minipage} 
            \\ \hline
            
            \centering
            Hide the Pain Harold
            &
            \begin{minipage}{.15\textwidth}
                \vspace{0.2cm}
                \begin{center}
                    \includegraphics[height= 1.5cm]{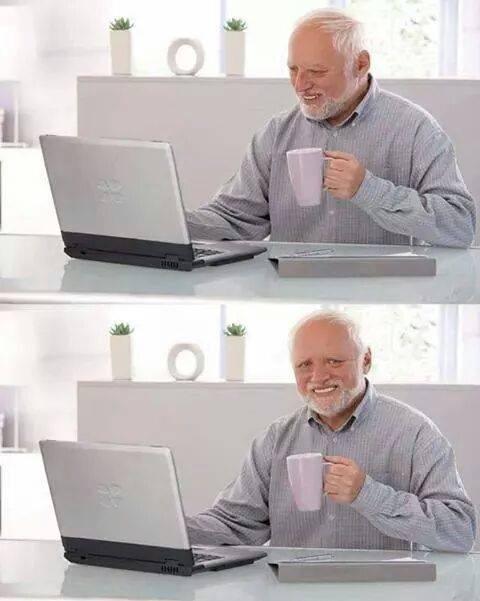}
                \end{center}
                \vspace{0.1cm}
            \end{minipage} 
            &
            \centering
            That Would Be Great
            &
            \begin{minipage}{.15\textwidth}
                \vspace{0.2cm}
                \begin{center}
                    \includegraphics[height= 1.5cm]{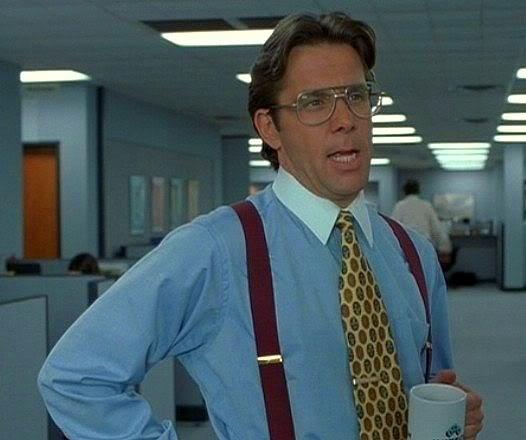}
                \end{center}
                \vspace{0.1cm}
            \end{minipage} 
            &
            \centering
            Grumpy Cat
            &
            \begin{minipage}{.15\textwidth}
                \vspace{0.2cm}
                \begin{center}
                    \includegraphics[height= 1.5cm]{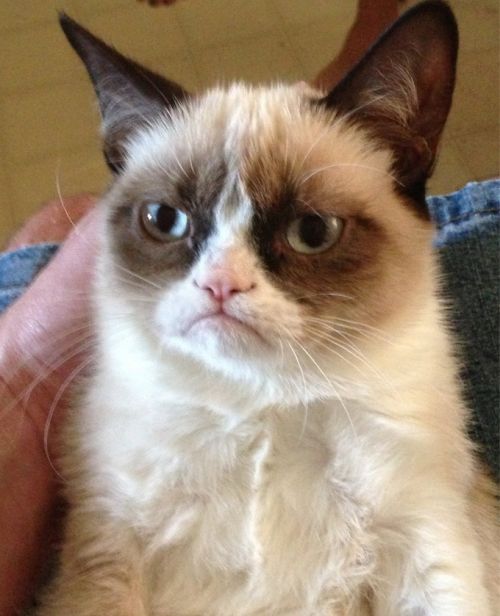}
                \end{center}
                \vspace{0.1cm}
            \end{minipage} 
            \\ \hline
        \end{tabular}
        \captionsetup{labelformat=empty}
        \caption{Meme templates and meme images from the meme caption dataset used in our experiments.}
    \end{center}
\end{table*}

\section{Sample Additional Images}

\begin{minipage}{\textwidth}
    \centering
    \includegraphics[width=\textwidth]{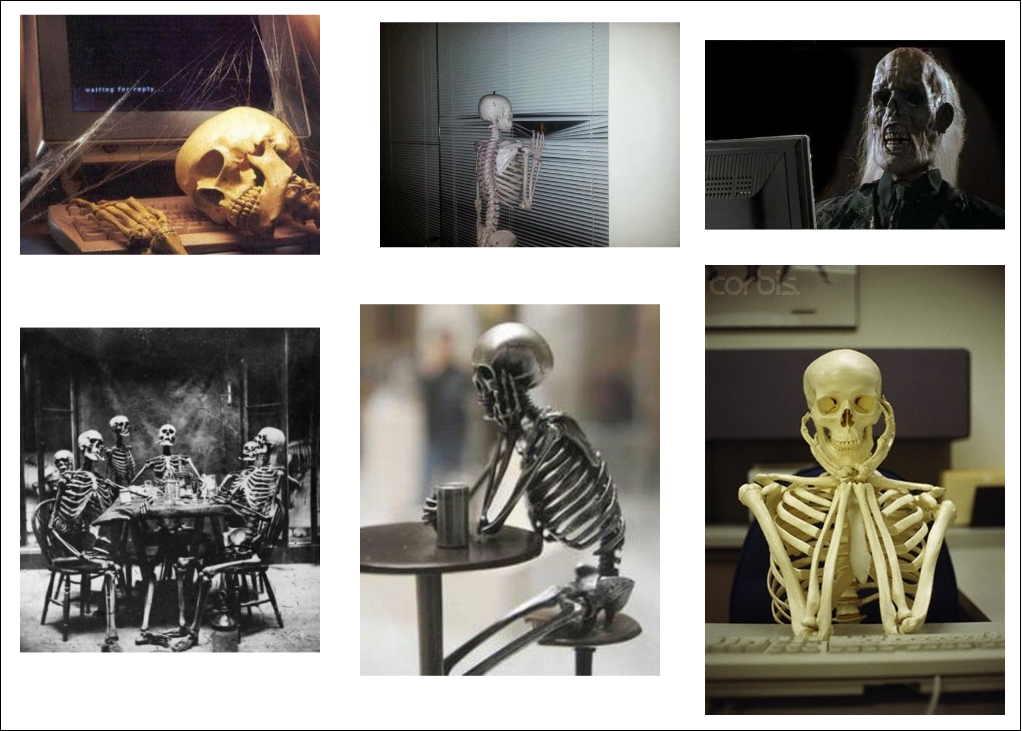}
    \captionsetup{labelformat=empty}
    \captionof{figure}{A Sample of additional images used for Waiting Skeleton meme template.}
\end{minipage}

\section{Amazon Mechanical Turk Questionnaire}

\begin{minipage}{\textwidth}
    \centering
    \includegraphics[width=\textwidth]{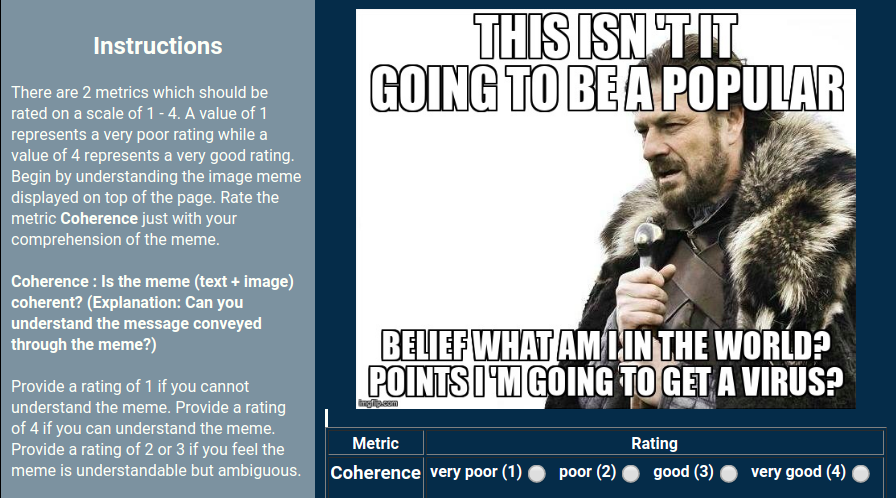}
    \captionsetup{labelformat=empty}
    \captionof{figure}{Sample AMT questionnaire - Coherence metric.}
\end{minipage}

\begin{minipage}{\textwidth}
    \centering
    \includegraphics[width=\textwidth]{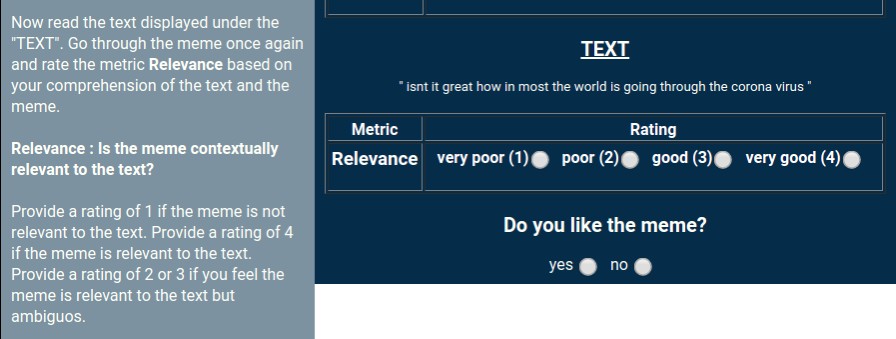}
    \captionsetup{labelformat=empty}
    \captionof{figure}{Sample AMT questionnaire - Relevance and User Likes metrics.}
\end{minipage}

\begin{minipage}{\textwidth}
    \centering
    \includegraphics[width=\textwidth]{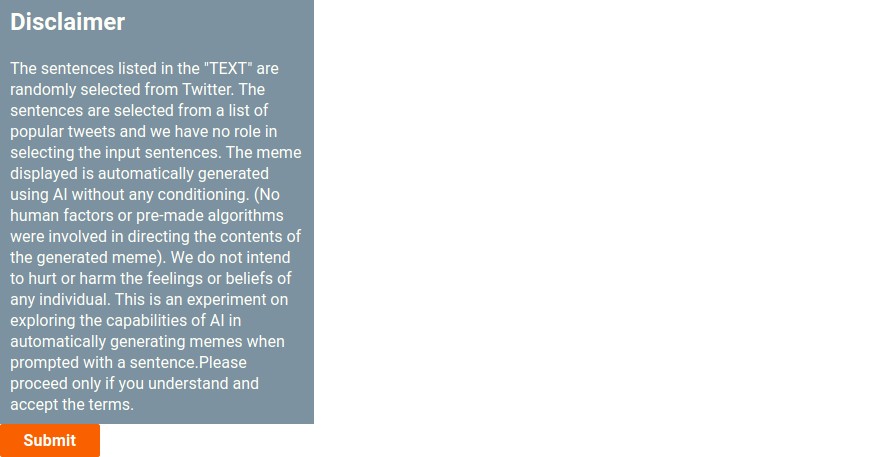}
    \captionsetup{labelformat=empty}
    \captionof{figure}{Sample AMT questionnaire - disclaimer.}
\end{minipage}

\section{memeBot Demo}

A live demo of the presented memeBot is available in the \color{darkblue}{\href{http://memebot-demo.s3-website-us-west-1.amazonaws.com/index.html}{website}}.

\end{document}